\documentclass[conference]{IEEEtran}
\IEEEoverridecommandlockouts

\usepackage{cite}
\usepackage{amsmath,amssymb,amsfonts}
\usepackage{graphicx}
\usepackage{textcomp}
\usepackage{xcolor}
\usepackage{multirow}
\usepackage{booktabs}
\usepackage{enumitem} 
\usepackage{algorithm}
\usepackage{algorithmic}
\usepackage{url}
\usepackage{balance}
\usepackage[colorlinks=true,
linkcolor=red,
citecolor=red,
urlcolor=red]{hyperref}
\usepackage{cleveref}
\newtheorem{myDef}{Definition}
\def\BibTeX{{\rm B\kern-.05em{\sc i\kern-.025em b}\kern-.08em
    T\kern-.1667em\lower.7ex\hbox{E}\kern-.125emX}}
\begin{document}

\title{When Correlations Mislead: Confounder-Aware Multi-View Urban Region Representation Learning -- Extended Version}

\author{\IEEEauthorblockN{Sean Bin Yang$^\dag$, Ying Sun$^\ddag$, Zongyi Xu$^\ddag$, Tung Kieu$^\dag$, \\ Jilin Hu$^\S$, Bin Yang$^\S$, Kristian Torp$^\dag$, Hua Lu$^\dag$, Torben Bach Pedersen$^\dag$}
	\IEEEauthorblockA{$^\dag$Aalborg University, Aalborg, Denmark \\
		$^\ddag$Chongqing University of Posts and Telecommunications, Chongqing, China\\
		$^\S$East China Normal University, Shanghai, China \\
		\textit{\{seany, tungkvt, torp, luhua, tbp\}@cs.aau.dk},
		\textit{\{sunying, xuzy\}@cqupt.edu.cn}, \\
		\textit{\{jlhu, byang\}@dase.ecnu.edu.cn},
	}
}

\maketitle

\begin{abstract}
Urban region representation learning commonly combines heterogeneous data sources, such as mobility flows, points of interest, and land-use information, to support tasks including mobility analysis, public safety forecasting, and service demand estimation. Existing multi-view methods typically improve region embeddings by strengthening interactions across views. However, such methods often overlook view-specific regional structures and may propagate correlations induced by shared latent factors, which can reduce the stability of downstream predictions.
To overcome this major limitation, we propose \texttt{CURE}, a confounder-aware framework for multi-view urban region representation learning. \texttt{CURE} first encodes each view with its regional graph structure, estimates a shared latent component, and then reduces its projected influence before cross-view interaction. A hierarchical graph-aware fusion module subsequently aggregates the residual view representations using local and global regional contexts.
Experiments on three real-world cities show that \texttt{CURE} improves predictive performance, remains robust under missing and noisy input views, and provides reliable cross-view integration through shared component separation and context-dependent view weighting.
	
\textbf{This is an extended version of '' When Correlations Mislead: Confounder-Aware Multi-View Urban Region Representation Learning'', to appear in ICDE 2027.}
\end{abstract}

\begin{IEEEkeywords}
Multi-view data integration, confounder-aware learning, urban region representation learning.
\end{IEEEkeywords}
\section{Introduction}

Urban region embeddings support a wide range of city analytics tasks, including mobility analysis~\cite{DBLP:journals/tkde/GongGLLZSLHW25,DBLP:journals/corr/abs-2511-20729,DBLP:journals/corr/abs-2106-09373,DBLP:conf/icde/YangGHYTJ22,DBLP:journals/corr/abs-2601-08482,DBLP:conf/kdd/YangHGYJ23,DBLP:journals/pvldb/PanWZY0CGWTDZYZ23}, location-based services~\cite{DBLP:conf/aaai/ZhouHCS023,DBLP:journals/tkde/GongWGLLZWZL24,DBLP:journals/tkde/GongGLLZSLHW25,DBLP:conf/CIKM/TFM,DBLP:journals/tkde/YangGY22,DBLP:conf/WWW/Path-LLM,DBLP:conf/KDD/MM-Path,DBLP:journals/corr/abs-2203-16110,DBLP:conf/aaai/HanYH26,DBLP:conf/icde/Yang020,DBLP:conf/kdd/Yang26,DBLP:conf/ijcai/YangGHT021,DBLP:journals/corr/abs-1907-04028}, and crime prediction~\cite{DBLP:conf/icde/LiHXXP22,DBLP:journals/tcss/ButtLAS25,DBLP:conf/aaai/WangLYSYS22,DBLP:conf/aaai/ZhaoFLT22,DBLP:journals/tkde/ZhaoLCZ23}. These tasks increasingly rely on heterogeneous observations, such as human mobility flows, points of interest (POIs), land-use records, and sensing data. Region representation learning provides a reusable way to encode such observations into low-dimensional vectors that preserve functional, spatial, and socioeconomic characteristics~\cite{DBLP:conf/kdd/SunCTK025,DBLP:conf/kdd/LiHCW023,DBLP:conf/ijcai/0004LLH20,DBLP:conf/icde/Sun0CFKT24,DBLP:journals/corr/abs-DGCPath,DBLP:journals/corr/abs-REFINE}.

Early methods primarily derive region embeddings from a dominant data source, such as mobility flows, POIs, building footprints, or geographic observations. Mobility-based approaches characterize regions through human transition patterns, while \texttt{RegionDCL}~\cite{DBLP:conf/kdd/LiHCW023} shows that building footprints, complemented by POIs, can also reveal useful regional semantics. A single view, however, captures only part of the information relevant to modeling an urban region.

Therefore, recent methods combine heterogeneous urban data. Multi-View Joint Graph Representation Learning~\cite{DBLP:conf/ijcai/0004LLH20} models mobility patterns and inherent region properties through cross-view information sharing and adaptive fusion. \texttt{HAFusion}~\cite{DBLP:conf/icde/Sun0CFKT24} focuses on attentive feature learning and fusion, while \texttt{CGAP}~\cite{DBLP:conf/ijcai/XuZ24} captures local and global graph structures through coarsened graph attention pooling. \texttt{FlexiReg}~\cite{DBLP:conf/kdd/SunCTK025} further adapts region embeddings to different spatial partitions and downstream tasks.
These methods rely heavily on observed dependencies across views and regions. However, such dependencies are not always reliable. Mobility, POI, and land-use views may be correlated because they reflect complementary urban semantics, but they may also respond to shared latent factors, such as population density, commercial intensity, transportation accessibility, and socioeconomic activity. As a result, directly strengthening these correlations can amplify shared biases and weaken view-specific signals.

\begin{figure*}[t]
	\centering
	\includegraphics[scale=0.9]{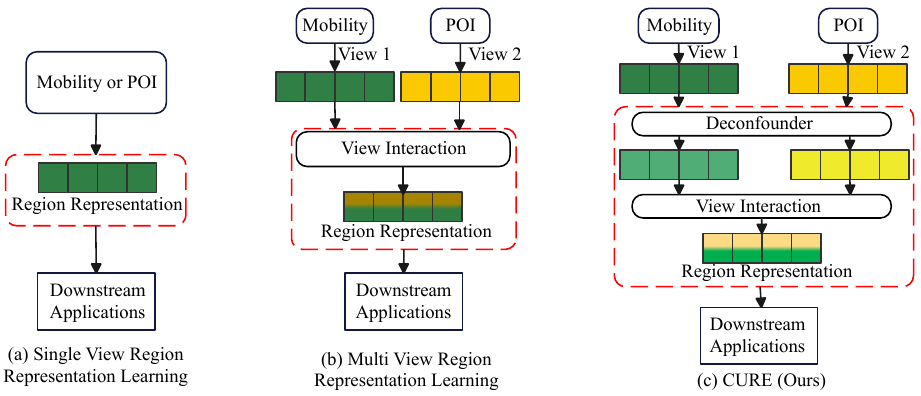}
	\caption{Comparison of region representation learning paradigms.
		(a) Single-view region representation learning constructs embeddings from one urban data source, which limits its ability to capture heterogeneous urban semantics.
		(b) Conventional multi-view region representation learning integrates multiple views through general cross-view interaction, but may directly propagate correlation-induced shared signals.
		(c) \texttt{CURE} explicitly models potential latent confounding factors and performs confounder-aware view interaction before fusion, enabling the learned region representations to emphasize stable and task-relevant view-specific information for downstream urban analytics.}
	\label{fig:example}
	\vspace{-10pt}
\end{figure*}

Fig.~\ref{fig:example} illustrates the distinction between existing paradigms and our motivation. Single-view methods learn region representations from one data source, limiting their ability to capture heterogeneous urban semantics. Conventional multi-view methods alleviate this limitation by integrating multiple views through general interaction and fusion. However, they typically assume that observed cross-view correlations are informative and reliable, while overlooking shared latent factors that may simultaneously influence different views. In urban data, factors such as population density, commercial intensity, transportation accessibility, and socioeconomic activity may induce correlations among mobility, POI, and land-use views without necessarily reflecting stable or task-relevant semantics \cite{DBLP:conf/aaai/0016DLJW23,DBLP:conf/nips/XiaLWLWZZ23,DBLP:conf/iclr/WangWDZ0PZL024,DBLP:journals/csur/GongZYBLX25,DBLP:conf/icde/Li0GCJZZFB24}. As a result, existing single-view and multi-view methods may encode biased shared signals or propagate spurious cross-view dependencies into learned region representations. 

%
In this work, we distinguish \emph{reliability} from \emph{robustness} in multi-view urban data integration. Reliability refers to whether the integrated representation reflects stable and task-relevant urban semantics rather than spurious cross-view correlations induced by shared latent factors. Robustness refers to whether the learned representation remains stable when the input views are incomplete or noisy. Under this distinction, reliability is challenged by correlation-driven view interaction and context-dependent view contribution, whereas robustness is challenged by data-quality variations such as missing or noisy views. This distinction motivates the following three challenges:

\noindent
\textbf{Challenge I: View-specific regional structure.}
Different urban views encode different types of regional dependencies. For example, mobility flows describe region-to-region movement connectivity, POI distributions capture functional similarity, and land-use records reflect planning-oriented spatial organization. Existing methods often treat these views mainly as heterogeneous feature sources to be fused, while overlooking the structural dependencies within each view. As a result, view-specific local patterns may be weakened before cross-view interaction, reducing the model's ability to preserve structured semantics from each data source. This also limits robustness, since the model may fail to exploit the remaining structured views when one view is incomplete or corrupted.

\noindent
\textbf{Challenge II: Correlation-driven cross-view interaction.}
Most multi-view methods strengthen observed dependencies across views, implicitly assuming that stronger cross-view correlations are beneficial. In urban data, however, different views may be correlated because they are influenced by shared latent factors, such as population density, commercial intensity, transportation accessibility, or socioeconomic activity. Directly propagating such correlations may amplify biased shared signals and introduce spurious cross-view dependencies. 
This creates a reliability risk: the learned representation may appear predictive by exploiting strong observed correlations, while these correlations may not correspond to stable or task-relevant urban semantics.

\noindent
\textbf{Challenge III: Context-dependent view contribution.}
Different urban views do not contribute equally across regions and tasks. Mobility signals may be highly informative in commercial centers but sparse or unstable in suburban regions. Land-use information may provide stable semantics in residential areas but may be less responsive to short-term activity patterns. POI distributions may capture functional composition but can be incomplete or unevenly updated. Existing fusion strategies mainly aim to increase representation strength, but they rarely model whether each view provides stable and task-relevant information under local and global regional contexts. This affects reliability, since a view that is informative in one region or task may introduce less relevant or unstable signals in another context, causing fixed or globally shared fusion strategies to overemphasize misleading view-specific information.

To address these challenges, we propose \texttt{CURE}, a \textbf{C}onfo\textbf{U}nder-awa\textbf{RE} framework for reliable multi-view urban region representation learning. \texttt{CURE} is designed following three principles. First, to address view-specific regional structure, \texttt{CURE} employs a graph-guided intra-view encoder that constructs and exploits a regional graph for each urban view. This enables mobility, POI, and land-use views to preserve their own structural semantics before cross-view interaction, providing more stable view-specific signals when some views are incomplete or noisy. Second, to address correlation-driven cross-view interaction, as shown in Fig. \ref{fig:example}(c), \texttt{CURE} introduces a confounder-aware inter-view interaction module that estimates a shared latent component across views and performs interaction in the residual representation space. This design reduces the direct propagation of shared latent correlations and encourages cross-view learning to focus on complementary view-specific information. Third, to address context-dependent view contribution, \texttt{CURE} develops a hierarchical graph-aware residual fusion module that aggregates residual view representations under both local and global graph contexts. By learning region-wise adaptive view weights, this module allows the model to emphasize views that provide more stable and task-relevant information under different urban structures and downstream tasks. Together, these designs enable \texttt{CURE} to improve predictive performance, provide reliable cross-view integration, and remain robust under incomplete or noisy urban observations.

Extensive experiments on three real-world cities demonstrate that \texttt{CURE} learns reliable region representations for urban analytics tasks, including check-in prediction, crime forecasting, and service call prediction. Compared with representative baselines, \texttt{CURE} consistently improves predictive performance across cities and tasks. Ablation studies verify the contribution of each component, while further analyses demonstrate both robustness under incomplete and noisy views and reliability in shared component separation and adaptive view weighting.

The main contributions are summarized as follows:
\begin{itemize}
	\item We propose \texttt{CURE}, a confounder-aware framework that integrates heterogeneous urban views into reliable region-level representations for downstream urban analytics.
	
	\item We design a graph-guided intra-view encoder that preserves view-specific regional structures and provides stable structured signals under incomplete or noisy views.
	
	\item We introduce a confounder-aware inter-view interaction module that estimates a shared latent component and performs cross-view learning in the residual space, reducing spurious dependencies induced by shared latent factors.
	
	\item We develop a hierarchical graph-aware residual fusion module that adaptively aggregates view-specific signals according to their context-dependent task relevance under local and global regional contexts.
	
	\item We conduct extensive experiments on three real-world cities and domain-specific analytics tasks, showing that \texttt{CURE} improves predictive performance, remains robust under incomplete and noisy views, and provides reliable cross-view integration.
\end{itemize}

\section{Related Work}

\subsection{Urban Region Representation Learning}

Urban region representation learning encodes urban observations into compact embeddings for tasks such as land-use classification, crime prediction, and population estimation. Early methods are often centered on a dominant data source. For instance, \texttt{MGFN}~\cite{DBLP:conf/ijcai/WuYFPZZ0W22} derives temporal mobility patterns from multiple mobility graphs, whereas \texttt{RegionDCL}~\cite{DBLP:conf/kdd/LiHCW023} primarily uses OpenStreetMap building footprints, complemented by POIs, to capture urban morphology and function. Such methods can model specific regional characteristics effectively, but they offer only a partial description of the urban environment.
Multi-view methods combine heterogeneous signals to obtain more comprehensive embeddings. Multi-View Joint Graph Representation Learning~\cite{DBLP:conf/ijcai/0004LLH20} introduces cross-view information sharing and adaptive fusion for mobility and regional attributes. \texttt{HREP}~\cite{DBLP:conf/aaai/ZhouHCS023} models intra-region features together with inter-region relations, while \texttt{CGAP}~\cite{DBLP:conf/ijcai/XuZ24} incorporates local graph structure and urban-level global information. \texttt{HAFusion}~\cite{DBLP:conf/icde/Sun0CFKT24} further studies attentive interaction within and across views. Other work improves the applicability of region embeddings across spatial partitions and tasks~\cite{DBLP:conf/kdd/LiHCW023,DBLP:conf/kdd/SunCTK025}.
Most of these methods treat stronger cross-view dependencies as useful signals. However, heterogeneous urban views may also be correlated through shared latent factors. Our work focuses on reducing the influence of such factors before cross-view interaction and fusion.

\subsection{Reliable Multi-View Urban Data Integration}

Multi-view learning integrates heterogeneous data sources that describe the same entities from complementary perspectives. In urban analytics, mobility flows, POI distributions, and land-use records respectively characterize movement connectivity, functional composition, and planning-oriented spatial structure.
Existing urban representation methods exploit heterogeneous information through cross-view information sharing and adaptive aggregation~\cite{DBLP:conf/ijcai/0004LLH20}, attentive feature learning and fusion~\cite{DBLP:conf/icde/Sun0CFKT24}, and graph-based integration of regional attributes and mobility contexts~\cite{DBLP:conf/ijcai/XuZ24}. Other methods employ contrastive learning to improve the generalizability of region embeddings across spatial partitions~\cite{DBLP:conf/kdd/LiHCW023}. Although effective, these approaches primarily focus on strengthening correlations across views or regions, and often treat the observed dependencies as reliable evidence.
This assumption may be problematic because heterogeneous urban views can be jointly influenced by shared latent factors, such as population density, transportation accessibility, commercial activity, and socioeconomic status. Directly reinforcing such correlations may amplify biased shared signals and obscure informative view-specific patterns. To address this issue, \texttt{CURE} estimates a shared latent component, reduces its projected
influence through soft residualization, and performs inter-view interaction
in the residual representation space. It further aggregates view-specific residual representations under local and global regional contexts, enabling context-dependent integration of stable and task-relevant urban signals.

\subsection{Confounder-Aware Representation Learning}

Confounding is a fundamental challenge in data-driven modeling, where observed dependencies may arise from latent factors rather than meaningful causal or semantic relationships \cite{DBLP:journals/natmi/CuiA22,DBLP:journals/corr/abs-2602-06240,DBLP:conf/cvpr/LvLLZLWL22,DBLP:conf/icde/ChuLR023,DBLP:conf/icde/ZhouWWYDY25,DBLP:journals/pvldb/LiuXAW25}. In representation learning, such confounders may cause embeddings to encode unstable or biased associations, thereby reducing robustness and generalization \cite{DBLP:journals/corr/abs-2505-17637,DBLP:conf/icml/AhujaMWB23,DBLP:conf/nips/AcarturkVST24,DBLP:conf/nips/RajendranBASR24}.
For example, stable learning \cite{DBLP:journals/natmi/CuiA22} incorporates ideas from causal inference to improve predictive modeling when stability, explainability, and fairness are important. Causal representation learning (CRL) \cite{DBLP:conf/icml/AhujaMWB23} further studies how interventional data can facilitate the identification of latent causal factors from low-level sensory observations. It shows that interventions provide geometric signatures that enable provable identification of latent factors, even without distributional or dependency assumptions.
This issue is particularly relevant to heterogeneous urban data, where latent domain factors may simultaneously influence mobility patterns, POI distributions, land-use characteristics, and downstream prediction targets. Although existing urban representation learning methods \cite{DBLP:conf/icde/Sun0CFKT24,DBLP:conf/ijcai/XuZ24,DBLP:conf/kdd/LiHCW023} capture rich correlations across heterogeneous views, they rarely account for latent confounding during cross-view interaction and fusion.
Unlike prior work, \texttt{CURE} introduces confounder-aware modeling into multi-view urban data integration. Specifically, it estimates a shared latent component, reduces its projected influence via soft residualization, and performs cross-view interaction only in the residual space. This design encourages complementary view-specific learning while suppressing spurious shared correlations, leading to more robust and generalizable urban representations.

\section{Preliminaries}

We first introduce the basic concepts used throughout the paper and then formulate the problem. The main notation is given in Table~\ref{tab:notation}.

\begin{table}[t]
	\scriptsize
	\centering
	\caption{Common notations.}
	\label{tab:notation}
	\begin{tabular}{ll}
		\toprule [1.5pt]
		\textbf{Notation} & \textbf{Description} \\
		\midrule
		$N$ & Number of urban regions \\
		$V$ & Number of urban data views \\
		$F_v$ & Feature dimension of the $v$-th view \\
		$d$ & Embedding dimension \\
		$\mathcal{R}=\{\mathcal{R}_1,\dots,\mathcal{R}_N\}$ & Set of urban regions \\
		$X^{(v)} \in \mathbb{R}^{N \times F_v}$ & Input feature matrix of the $v$-th view \\
		$A^{(v)} \in \mathbb{R}^{N \times N}$ & View-specific regional graph of the $v$-th view \\
		$A^{\text{local}} \in \mathbb{R}^{N \times N}$ & Local regional graph used in fusion \\
		$A^{\text{global}} \in \mathbb{R}^{N \times N}$ & Global functional-similarity graph used in fusion \\
		$H^{(v)} \in \mathbb{R}^{N \times d}$ & Intra-view representation of the $v$-th view \\
		$C \in \mathbb{R}^{N \times d}$ & Estimated shared latent component across views \\
		$R^{(v)} \in \mathbb{R}^{N \times d}$  & Residual representation of the $v$-th view \\
		$\alpha^{(v)} \in \mathbb{R}^{N \times 1}$ & Adaptive fusion weight of the $v$-th view \\
		$H \in \mathbb{R}^{N \times d}$ & Final urban region embedding \\
		$\lambda$ & Weight of the decorrelation regularizer \\
		\bottomrule[1.5pt]
	\end{tabular}
	\vspace{-10pt}
\end{table}

\subsection{Definitions}

\begin{myDef}[Urban Regions]
	Given a study area $\mathcal{A}$, let $\mathcal{R}=\{\mathcal{R}_1,\dots,\mathcal{R}_N\}$ denote a set of non-overlapping urban regions that partition $\mathcal{A}$, where $N$ is the number of regions. Formally, these regions satisfy $\mathcal{R}_i \cap \mathcal{R}_j=\emptyset$ for $i \neq j$ and $\bigcup_{i=1}^{N}\mathcal{R}_i=\mathcal{A}$. Each region $\mathcal{R}_i$ corresponds to a geographically bounded spatial unit, such as a grid cell, census tract, or administrative zone. In this work, urban regions serve as the basic analytical units for integrating heterogeneous urban data, constructing region-level representations, and supporting downstream urban analytics tasks.
\end{myDef}

\begin{myDef}[Mobility, POI, and Land-Use Views]
	The mobility view characterizes movement connectivity among regions and is represented by a region-level mobility matrix $X^{(1)} \in \mathbb{R}^{N \times N}$. Each entry $X_{ij}^{(1)}$ records the flow volume from region $\mathcal{R}_i$ to region $\mathcal{R}_j$ during a given observation period, such as taxi trips, human mobility flows, or commuting records. This view reflects dynamic spatial interactions and transportation-related dependencies between regions.
	
	The POI view describes the functional composition of each region based on the distribution of POI categories. It is represented by a feature matrix $X^{(2)} \in \mathbb{R}^{N \times F_2}$, where each feature dimension corresponds to a POI category or a derived functional attribute. The POI view captures region-level urban functions, such as commercial activity, residential services, education, healthcare, recreation, and transportation facilities.
	
	The land-use view reflects the planning-oriented spatial structure of each region and is represented by a feature matrix $X^{(3)} \in \mathbb{R}^{N \times F_3}$. Each row summarizes the land-use composition of a region, such as residential, commercial, industrial, public service, green space, or transportation-related land-use types. Compared with the POI view, which reflects fine-grained functional facilities, the land-use view provides a more structural and planning-level description of urban space.
\end{myDef}

\begin{myDef}[Multi-View Urban Features]
	Given the region set $\mathcal{R}$, each region is associated with multiple heterogeneous urban data views that describe different aspects of urban semantics. We denote the collection of multi-view features as
	\begin{equation}
		\mathcal{X}=\{X^{(1)},X^{(2)},\dots,X^{(V)}\},
	\end{equation}
	where $V$ is the number of views and $X^{(v)} \in \mathbb{R}^{N \times F_v}$ denotes the feature matrix of the $v$-th view. The $i$-th row $X_i^{(v)} \in \mathbb{R}^{F_v}$ represents the feature vector of region $\mathcal{R}_i$ under view $v$, and $F_v$ is the corresponding feature dimensionality. Different views may have different feature dimensionalities, data distributions, and semantic meanings. In this work, we consider three representative urban views: mobility, points of interest (POIs), and land-use information. Nevertheless, our method naturally extends to more views.
\end{myDef}

\begin{figure*}[t]
	\centering
	\includegraphics[scale=0.44]{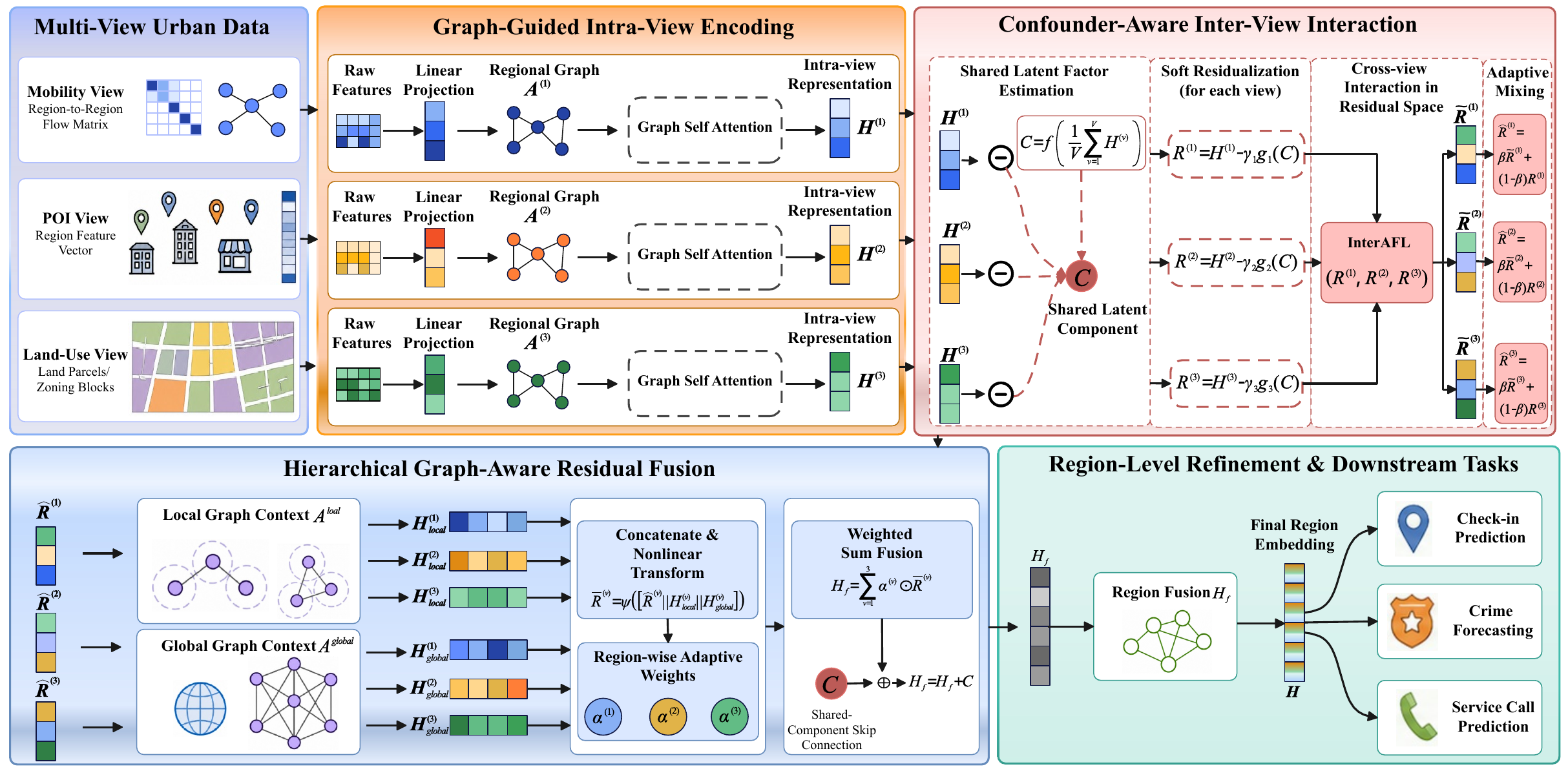}
	\caption{Overview of \texttt{CURE}. \texttt{CURE} encodes mobility, POI,
		and land-use views with graph-guided intra-view encoders, estimates a shared
		latent component and reduces its projected influence before cross-view
		interaction, and adaptively fuses residual view representations under local
		and global graph contexts. The resulting region embeddings are used for
		downstream tasks such as check-in prediction, crime forecasting, and service
		call prediction.}
	\label{fig:overview}
	\vspace{-10pt}
\end{figure*}

\subsection{Problem Definition}

Given the region set $\mathcal{R}$ and the associated multi-view urban features $\mathcal{X}=\{X^{(1)},X^{(2)},\dots,X^{(V)}\}$, the objective of reliable multi-view urban region representation learning is to learn a unified embedding matrix
\begin{equation}
	H \in \mathbb{R}^{N \times d},
\end{equation}
where $d$ is the embedding dimensionality and the $i$-th row $H_i \in \mathbb{R}^{d}$ denotes the learned representation of region $\mathcal{R}_i$. In this work, $H$ is learned by optimizing a structure-preserving objective that reconstructs pairwise regional similarities from mobility, POI, and land-use views, together with a decorrelation regularizer that reduces the dependence between residual view representations and the estimated shared latent component. Therefore, a desirable embedding should preserve view-specific regional structures, suppress misleading shared correlations, and support downstream urban analytics tasks, such as check-in prediction, crime forecasting, and service call prediction.

\section{Methodology}

We propose \texttt{CURE}, a confounder-aware framework for reliable multi-view urban region representation learning. Given heterogeneous urban views, including mobility, POIs, and land-use information, \texttt{CURE} aims to learn unified region embeddings that preserve view-specific semantics while mitigating spurious cross-view dependencies caused by shared latent factors. 

\subsection{Framework Overview}

Let $\mathcal{V}=\{v_1, v_2, \dots, v_V\}$ denote the set of urban views. For each view $v \in \mathcal{V}$, we denote its feature matrix by $X^{(v)} \in \mathbb{R}^{N \times F_v}$, where $N$ is the number of urban regions and $F_v$ is the feature dimensionality of view $v$. In this work, we consider three representative urban views, namely mobility, POIs, and land-use. The objective is to learn a unified region embedding matrix $H \in \mathbb{R}^{N \times d}$ that can support downstream urban analytics tasks.

As illustrated in Fig.~\ref{fig:overview}, the framework consists of four key components: graph-guided intra-view encoding, confounder-aware inter-view interaction, hierarchical graph-aware residual fusion, and region-level refinement. Specifically, \texttt{CURE} first preserves view-specific regional structures through graph-guided intra-view encoding, which provides stable structured signals under incomplete or noisy views. It then estimates a shared latent component and performs inter-view interaction in the residual space, improving reliability by reducing spurious cross-view dependencies induced by shared latent factors. Next, hierarchical graph-aware residual fusion adaptively aggregates residual view representations under local and global graph contexts, improving context-dependent reliability and robustness under input perturbations. Finally, region-level refinement captures higher-order dependencies and produces the unified region embedding $H$.

\subsection{Graph-Guided Intra-View Encoding}

Different urban views encode different regional structures that may exhibit dependencies across views. To preserve such view-specific dependencies, we associate each view with a regional graph $A^{(v)} \in \mathbb{R}^{N \times N}$ and encode it independently.

For view $v$, we first project its raw features into a shared latent space:
\begin{equation}
	Z_0^{(v)} = \phi_v(X^{(v)}),
\end{equation}
where $\phi_v(\cdot)$ is a learnable projection and $Z_0^{(v)} \in \mathbb{R}^{N \times d}$.

We first add self-loops and row-normalize the view-specific adjacency matrix:
\begin{equation}
	\tilde{A}^{(v)}
	=
	\left(D^{(v)}\right)^{-1}
	\left(A^{(v)} + I\right),
\end{equation}
where $D^{(v)}$ is the corresponding degree matrix.

At the $l$-th encoder block, graph propagation is performed as:
\begin{equation}
	G_l^{(v)}
	=
	\tilde{A}^{(v)} Z_l^{(v)} W_{g,l}^{(v)}.
\end{equation}

The graph-enhanced representation is obtained through a residual connection
and layer normalization:
\begin{equation}
	Z_{l}^{\prime(v)}
	=
	\operatorname{LayerNorm}
	\left(
	Z_l^{(v)} + \operatorname{Dropout}(G_l^{(v)})
	\right).
\end{equation}

Multi-head self-attention (MHA) and a feed-forward network (FFN) are then applied:
\begin{equation}
	S_l^{(v)}
	=
	\operatorname{MHA}
	\left(
	Z_{l}^{\prime(v)},
	Z_{l}^{\prime(v)},
	Z_{l}^{\prime(v)}
	\right),
\end{equation}
\begin{equation}
	Z_{l}^{\prime\prime(v)}
	=
	\operatorname{LayerNorm}
	\left(
	Z_{l}^{\prime(v)}
	+
	\operatorname{Dropout}(S_l^{(v)})
	\right),
\end{equation}
\begin{equation}
	Z_{l+1}^{(v)}
	=
	\operatorname{LayerNorm}
	\left(
	Z_{l}^{\prime\prime(v)}
	+
	\operatorname{Dropout}
	\left(
	\operatorname{FFN}(Z_{l}^{\prime\prime(v)})
	\right)
	\right).
\end{equation}
After stacking $L_G$ graph-guided encoder blocks, we obtain the intra-view
representation:
\begin{equation}
	H^{(v)}
	=
	\phi_{\mathrm{out}}^{(v)}
	\left(
	Z_{L_G}^{(v)}
	\right),
\end{equation}
where $H^{(v)} \in \mathbb{R}^{N \times d}$ and
$\phi_{\mathrm{out}}^{(v)}(\cdot)$ is a learnable output transformation.

\subsection{Confounder-Aware Inter-View Interaction}

Heterogeneous urban views often share latent factors, which may induce spurious cross-view correlations. The intuition is that signals consistently shared across views may capture common urban background factors, such as population density, commercial intensity, or transportation accessibility, rather than complementary task-relevant semantics. Therefore, instead of directly interacting the original view representations, \texttt{CURE} estimates a shared latent component as a learnable proxy for such cross-view commonality.

Given this component, \texttt{CURE} reduces its projected influence from each view through soft residualization. The resulting residual representations retain more view-specific information while suppressing correlations already explained by the shared component. Cross-view interaction is then performed in the residual space, encouraging the model to focus on complementary view-specific semantics:

\begin{equation}
	C = f\!\left(\frac{1}{V}\sum_{v=1}^{V} H^{(v)}\right),
\end{equation}
where $f(\cdot)$ is a learnable nonlinear mapping and $C \in \mathbb{R}^{N \times d}$ denotes the shared latent component.

For each view, we reduce the projected influence of $C$ through soft residualization:
\begin{equation}
	R^{(v)} = H^{(v)} - \gamma_v g_v(C),
\end{equation}
where $g_v(\cdot)$ is a view-specific projection and $\gamma_v=\sigma(\theta_v)$ is a learnable coefficient constrained to $(0,1)$. This design reduces the influence of dominant shared signals while avoiding over-removal of useful common semantics.

The residual representations are organized into a region-wise view sequence:
\begin{equation}
	Q_0
	=
	\operatorname{Stack}
	\left(
	R^{(1)}, \dots, R^{(V)}
	\right)
	\in \mathbb{R}^{N \times V \times d}.
\end{equation}
In particular, we denote the inter-view attentive feature learning module as \texttt{InterAFL}, which models dependencies among different residual views for each region. 

At the $l$-th inter-view block, the residual representations are projected
into an intermediate attention space:
\begin{equation}
	E_l = Q_l W_{k,l},
\end{equation}
where $W_{k,l}\in\mathbb{R}^{d\times d_m}$.

The attention features are normalized across views:
\begin{equation}
	A_{l,n,v,s}
	=
	\frac{
		\exp(E_{l,n,v,s})
	}{
		\sum_{u=1}^{V}\exp(E_{l,n,u,s})
	}.
\end{equation}
where $n$, $v$, and $s$ index the region, view, and intermediate attention
dimension, respectively.

The attention features are further normalized along the intermediate dimension:
\begin{equation}
	\bar{A}_{l,n,v,s}
	=
	\frac{
		A_{l,n,v,s}
	}{
		\sum_{j=1}^{d_m} A_{l,n,v,j}
	}.
\end{equation}

The block output is:
\begin{equation}
	Q_{l+1} = \bar{A}_l W_{o,l},
\end{equation}
where $W_{o,l}\in\mathbb{R}^{d_m\times d}$ is a learnable output projection.

After stacking $L_I$ blocks, the interacted residual representations are:
\begin{equation}
	\left[
	\tilde{R}^{(1)},\dots,\tilde{R}^{(V)}
	\right]
	=
	\operatorname{MLP}(Q_{L_I}).
\end{equation}

We further combine the interacted and original residual representations:
\begin{equation}
	\hat{R}^{(v)} = \beta \tilde{R}^{(v)} + (1-\beta) R^{(v)},
\end{equation}
where $\beta=\sigma(\theta_\beta)$ is a learnable mixing coefficient constrained to $(0,1)$. This encourages cross-view learning to focus on complementary view-specific information.

\subsection{Hierarchical Graph-Aware Residual Fusion}

The contribution of each view to reliable integration may vary across regions and structural contexts. Therefore, we fuse residual view representations with local and global graph contexts. Let $A^{\text{local}} \in \mathbb{R}^{N \times N}$ denote a local regional graph and $A^{\text{global}} \in \mathbb{R}^{N \times N}$ denote a global functional-similarity graph. For each view $v$, we compute:
\begin{equation}
	H_{\text{local}}^{(v)} = A^{\text{local}} \hat{R}^{(v)} W_l,
\end{equation}
\begin{equation}
	H_{\text{global}}^{(v)} = A^{\text{global}} \hat{R}^{(v)} W_h,
\end{equation}
where $W_l,W_h \in \mathbb{R}^{d \times d}$ are learnable matrices.

The residual representation and its graph-enhanced contexts are combined as:
\begin{equation}
	\bar{R}^{(v)} =
	\psi\!\left([\hat{R}^{(v)} \| H_{\text{local}}^{(v)} \| H_{\text{global}}^{(v)}]\right),
\end{equation}
where $\psi(\cdot)$ is a nonlinear transformation and $\|$ denotes concatenation.

Region-wise fusion weights are computed by:
\begin{equation}
	e^{(v)} = s(\bar{R}^{(v)}),
\end{equation}
\begin{equation}
	\alpha^{(v)} =
	\frac{\exp(e^{(v)})}
	{\sum_{u=1}^{V}\exp(e^{(u)})},
\end{equation}
where $s(\cdot)$ is a learnable scoring function and $\alpha^{(v)} \in \mathbb{R}^{N \times 1}$. The fused representation is given as follows:
\begin{equation}
	H_f
	=
	\sum_{v=1}^{V}
	\alpha^{(v)} \odot \bar{R}^{(v)}
	+
	C.
\end{equation}

This fusion module adaptively emphasizes view-specific signals that are more stable and task-relevant under local and global regional structures.

\subsection{Region-Level Refinement}

The fused representation is further refined to capture higher-order dependencies among regions:
\begin{equation}
	H = \mathrm{RegionFusion}(H_f),
\end{equation}
where $\mathrm{RegionFusion}(\cdot)$ denotes a Transformer-style region
refinement module composed of multi-head self-attention, residual connections,
layer normalization, and a feed-forward network.

\subsection{Training of \texttt{CURE}}

We train \texttt{CURE} with a multi-objective loss that preserves the
region-level similarity structures encoded by mobility, POI, and land-use
features. Let $x_i^{s}$ and $x_i^{t}$ denote the outgoing and incoming
mobility flow vectors of region $i$, respectively, and let $x_i^{p}$ and
$x_i^{l}$ denote its POI and land-use feature vectors. We first map the
unified region embedding $H$ into feature-oriented representations:
\begin{equation}
	H^{q} = \phi_q(H), \qquad q \in \{s,t,p,l\},
\end{equation}
where $\phi_q(\cdot)$ is a learnable decoder and
$H^{q}=\{h_1^{q},\dots,h_N^{q}\}$.

Specifically, for each feature type \(q \in \{s,t,p,l\}\), we define
\begin{equation}
	\mathcal{L}_{q}
	=
	\frac{1}{N^2}
	\sum_{i=1}^{N}
	\sum_{j=1}^{N}
	\left|
	S_{ij}^{q}
	-
	\operatorname{sim}
	\left(
	h_i^{q}, h_j^{q}
	\right)
	\right|,
\end{equation}
where \(s\) and \(t\) denote outgoing and incoming mobility structures, while \(p\) and \(l\) denote POI and land-use structures, respectively. \(S_{ij}^{q}=\operatorname{sim}(x_i^{q},x_j^{q})\) is the pairwise similarity computed from the original feature type \(q\), and \(h_i^{q}\) is the corresponding decoded representation of region $\mathcal{R}_i$.

The mobility objective jointly preserves outgoing and incoming flow
structures:
\begin{equation}
	\mathcal{L}_{\text{mob}}
	=
	\mathcal{L}_{s}
	+
	\mathcal{L}_{t}.
\end{equation}
The POI and land-use objectives are defined as
$\mathcal{L}_{\text{poi}}=\mathcal{L}_{p}$ and
$\mathcal{L}_{\text{land}}=\mathcal{L}_{l}$, respectively. The main
training objective is:
\begin{equation}
	\mathcal{L}_{\text{main}}
	=
	\mathcal{L}_{\text{mob}}
	+
	\mathcal{L}_{\text{poi}}
	+
	\mathcal{L}_{\text{land}}.
\end{equation}

To reduce the dependence between the shared latent component and the residual view representations, we introduce a decorrelation regularizer:
\begin{equation}
	\mathcal{L}_{\text{decor}}
	=
	\frac{1}{V}
	\sum_{v=1}^{V}
	\left\|
	\mathrm{Corr}\!\left(R^{(v)}, C\right)
	\right\|_F^2,
\end{equation}
where $\mathrm{Corr}\!\left(R^{(v)}, C\right)\in\mathbb{R}^{d\times d}$
denotes the dimension-wise Pearson correlation matrix between
$R^{(v)}$ and $C$, computed across the $N$ urban regions.

The final objective is:
\begin{equation}
	\mathcal{L}
	=
	\mathcal{L}_{\text{main}}
	+
	\lambda \mathcal{L}_{\text{decor}},
\end{equation}
where $\lambda$ controls the strength of the decorrelation regularizer. In practice, we gradually introduce $\mathcal{L}_{\text{decor}}$ during training to stabilize optimization. Through joint optimization, \texttt{CURE} learns region representations that preserve view-specific structures, reduce the influence of shared latent factors, and support reliable urban analytics.

\section{Experiments}

\subsection{Experimental Settings}

\begin{table}[t]
	\scriptsize
	\centering
	
	\renewcommand{\arraystretch}{1.2}  %
	\setlength{\tabcolsep}{0.4pt} %
	\caption{Dataset statistics}
	\begin{tabular}{l|l|l|l}
		\toprule[1.5pt]
		& \textbf{NY}      & \textbf{Chi}      & \textbf{SF}       \\ \hline
		{\#Regions}              & 180               & 77                & 175               \\ \hline
		{\#POIs}                 & 24,496            & 57,891            & 28,578            \\ \hline
		{\#POI categories}       & 26                & 26                & 26                \\ \hline
		{\#Land use categories}  & 11                & 12                & 23                \\ \hline
		{\#Taxi trips}           & 10,953,879        & 3,381,807         & 357,749           \\ \hline
		{(Collection period)} & 2015/06 - 2015/07 & 2021/01 - 2022/01 & 2008/05 - 2008/06 \\ \hline
		{\#Crime records}        & 35,335            & 18,200            & 48,489            \\ \hline
		{(Collection period)} & unknown           & 2022/12 - 2022/12 & 2011/01 - 2022/12 \\ \hline
		{\#Check-ins}            & 106,902           & 167,232           & 87,750            \\ \hline
		{(Collection period)} & 2012/04 - 2013/09 & 2012/04 - 2013/09 & 2012/04 - 2013/09 \\ \hline
		{\#Service calls}        & 516,187           & 24,350            & 34,385            \\ \hline
		{(Collection period)} & 2023/01 - 2023/03 & 2022/12 - 2022/12 & 2022/01 - 2022/12 \\ \bottomrule[1.5pt]
	\end{tabular}
	\label{tab:dataset}
	\vspace{-10pt}
\end{table}

\begin{table*}[t]
	\centering
	\small
	\setlength{\tabcolsep}{2pt} %
	\renewcommand{\arraystretch}{1.9}  %
	\caption{Overall Performance on three datasets. '$\downarrow$' denotes that smaller values are preferred, and '$\uparrow$' denotes that larger values are preferred. We bold the best performance in \textcolor{red}{red} and underline the second-best results in \textcolor{blue}{blue} for each dataset.}
	\begin{tabular}{l|c|ccc|ccc|ccc}
		\toprule[1.5pt]
		\multirow{2}{*}{\textbf{}}             & \multicolumn{1}{c|}{\multirow{2}{*}{\textbf{Method}}} & \multicolumn{3}{c|}{\textbf{NY}}                                            & \multicolumn{3}{c|}{\textbf{Chi}}                                                  & \multicolumn{3}{c}{\textbf{SF}}                                            \\ \cline{3-11} 
		& \multicolumn{1}{c|}{}                                  & \multicolumn{1}{c|}{\textbf{MAE} ($\downarrow$)} & \multicolumn{1}{c|}{\textbf{RMSE} ($\downarrow$)} & \multicolumn{1}{c|}{\textbf{$R^2$} ($\uparrow$)}  & \multicolumn{1}{c|}{\textbf{MAE} ($\downarrow$)} & \multicolumn{1}{c|}{\textbf{RMSE} ($\downarrow$)} &\multicolumn{1}{c|}{ \textbf{$R^2$} ($\uparrow$)}   & \multicolumn{1}{c|}{\textbf{MAE} ($\downarrow$)} & \multicolumn{1}{c|}{\textbf{RMSE} ($\downarrow$)} & \multicolumn{1}{c}{\textbf{$R^2$} ($\uparrow$)}   \\ \hline \hline
		\multirow{6}{*}{\textbf{\rotatebox{90}{Check-in}}}     & \texttt{MVURE}                                         & \multicolumn{1}{l|}{306.7 ± 8.20}   & \multicolumn{1}{l|}{499.6 ± 12.9}    & 0.627 ± 0.019 & \multicolumn{1}{l|}{1693 ± 74}    & \multicolumn{1}{l|}{3171 ± 128}      & 0.656 ± 0.029 & \multicolumn{1}{l|}{346.8 ± 8.7}    & \multicolumn{1}{l|}{659.3 ± 15.7}  & 0.562 ± 0.021 \\ \cline{2-11} 
		& \texttt{MGFN}                                          & \multicolumn{1}{l|}{292.6 ± 17.1}   & \multicolumn{1}{l|}{451.8 ± 28.1}  & 0.690 ± 0.040 & \multicolumn{1}{l|}{1281 ± 41}    & \multicolumn{1}{l|}{2276 ± 86}    & 0.817 ± 0.011 & \multicolumn{1}{l|}{310.8 ± 9.1}  & \multicolumn{1}{l|}{542.1 ± 17.6}  & 0.708 ± 0.010 \\ \cline{2-11} 
		& \texttt{RDCL}                                     & \multicolumn{1}{l|}{371.2 ± 10.3}   & \multicolumn{1}{l|}{495.5 ± 15.9}  & 0.471 ± 0.023 & \multicolumn{1}{l|}{2427 ± 123}   & \multicolumn{1}{l|}{4184 ± 136}     & 0.402 ± 0.042 & \multicolumn{1}{l|}{398.8  ± 9.9  }  & \multicolumn{1}{l|}{748.1 ± 17.8}  & 0.437 ± 0.024 \\ \cline{2-11} 
		& \texttt{HREP}                                          & \multicolumn{1}{l|}{276.3 ± 11.7}   & \multicolumn{1}{l|}{448.2 ± 17.1}  & 0.703 ± 0.021 & \multicolumn{1}{l|}{1679 ± 71}    & \multicolumn{1}{l|}{3135 ± 79}    & 0.664 ± 0.017 & \multicolumn{1}{l|}{330.9 ± 9.3}  & \multicolumn{1}{l|}{606.7 ± 25.8}  & 0.629 ± 0.032 \\ \cline{2-11} 
		& \texttt{HAF}                                      & \multicolumn{1}{l|}{\textcolor{blue}{\underline{202.8 ± 7.2}}}    & \multicolumn{1}{l|}{\textcolor{blue}{\underline{322.8 ± 12.6}}}  & \textcolor{blue}{\underline{0.844 ± 0.012}} & \multicolumn{1}{l|}{\textcolor{blue}{\underline{929 ± 62}}}     & \multicolumn{1}{l|}{\textcolor{blue}{\underline{1947 ± 75}}}     &\textcolor{blue}{\underline{ 0.870 ± 0.010}} & \multicolumn{1}{l|}{\textcolor{blue}{\underline{233.1 ± 9.5 }}}  & \multicolumn{1}{l|}{\textcolor{blue}{\underline{429.6 ± 28.1}}}  & \textcolor{blue}{\underline{0.813 ± 0.024}} \\ \cline{2-11} 
		& \texttt{CURE}                                     & \multicolumn{1}{l|}{\textcolor{red}{\textbf{186.7 ± 2.2}}}             & \multicolumn{1}{l|}{\textcolor{red}{\textbf{289.0 ± 4.6}}}              & \textcolor{red}{\textbf{0.875 ± 0.004}}              & \multicolumn{1}{l|}{\textcolor{red}{\textbf{762 ± 56}}}             & \multicolumn{1}{l|}{\textcolor{red}{\textbf{1569 ± 19}}}              & \textcolor{red}{\textbf{0.916 ± 0.011}}              & \multicolumn{1}{l|}{\textcolor{red}{\textbf{215.8 ± 5.5}}}             & \multicolumn{1}{l|}{\textcolor{red}{\textbf{373.4 ± 6.0}}}              &\textcolor{red}{\textbf{0.860 ± 0.004}}               \\ \hline \hline 
		\multirow{6}{*}{\textbf{\rotatebox{90}{Crime}}}        & \texttt{MVURE}                                         & \multicolumn{1}{l|}{67.9 ± 1.1}   & \multicolumn{1}{l|}{93.8 ± 1.9}    & 0.591 ± 0.016 & \multicolumn{1}{l|}{100.4 ± 6.6}  & \multicolumn{1}{l|}{129.2 ± 7.3}   & 0.461 ± 0.062 & \multicolumn{1}{l|}{130.3 ± 1.7}  & \multicolumn{1}{l|}{201.7 ± 3.2}   & 0.594 ± 0.013 \\ \cline{2-11} 
		& \texttt{MGFN}                                          & \multicolumn{1}{l|}{70.2 ± 2.3}   & \multicolumn{1}{l|}{89.6 ± 2.5}    & 0.630 ± 0.020 & \multicolumn{1}{l|}{107.4 ± 5.4}  & \multicolumn{1}{l|}{137.9 ± 5.2}   & 0.386 ± 0.047 & \multicolumn{1}{l|}{128.4 ± 3.3}  & \multicolumn{1}{l|}{199.9 ± 4.3}   & 0.601 ± 0.017 \\ \cline{2-11} 
		& \texttt{RDCL}                                     & \multicolumn{1}{l|}{98.7 ± 3.1}   & \multicolumn{1}{l|}{127.9 ± 5.2}   & 0.251 ± 0.026 & \multicolumn{1}{l|}{121.7 ± 4.8}  & \multicolumn{1}{l|}{159.6 ± 6.3}   & 0.179 ± 0.053 & \multicolumn{1}{l|}{156.3 ± 2.1}  & \multicolumn{1}{l|}{242.3 ± 4.6}   & 0.413 ± 0.021 \\ \cline{2-11} 
		& \texttt{HREP}                                          & \multicolumn{1}{l|}{62.8 ± 2.1}   & \multicolumn{1}{l|}{83.1 ± 2.3}    & 0.680 ± 0.014 & \multicolumn{1}{l|}{88.3 ± 6.4}   & \multicolumn{1}{l|}{114.4 ± 5.5}   & 0.578 ± 0.041 & \multicolumn{1}{l|}{124.4 ± 2.3}  & \multicolumn{1}{l|}{196.9 ± 3.9}   & 0.612 ± 0.014 \\ \cline{2-11} 
		& \texttt{HAF}                                      & \multicolumn{1}{l|}{\textcolor{blue}{\underline{56.1 ± 1.3}}}   & \multicolumn{1}{l|}{\textcolor{blue}{\underline{76.1 ± 2.2}}}    & \textcolor{blue}{\underline{0.734 ± 0.015}} & \multicolumn{1}{l|}{\textcolor{blue}{\underline{77.8 ± 3.6}}}   & \multicolumn{1}{l|}{\textcolor{blue}{\underline{107.1 ± 5.4}}}   & \textcolor{blue}{\underline{0.631 ± 0.036}} & \multicolumn{1}{l|}{\textcolor{blue}{\underline{101.5 ± 3.3}}}  & \multicolumn{1}{l|}{\textcolor{blue}{\underline{178.4 ± 3.6}}}   & \textcolor{blue}{\underline{0.682 ± 0.013}} \\ \cline{2-11} 
		& \texttt{CURE}                                     & \multicolumn{1}{l|}{\textcolor{red}{\textbf{53.2 ± 1.5}}}             & \multicolumn{1}{l|}{\textcolor{red}{\textbf{73.1 ± 0.9}}}              &\textcolor{red}{\textbf{0.755 ± 0.006}}               & \multicolumn{1}{l|}{\textcolor{red}{\textbf{69.6 ± 3.2}}}             & \multicolumn{1}{l|}{\textcolor{red}{\textbf{91.2 ± 0.9}}}              &\textcolor{red}{\textbf{0.732 ± 0.005}}               & \multicolumn{1}{l|}{\textcolor{red}{\textbf{99.1 ± 0.6}}}             & \multicolumn{1}{l|}{\textcolor{red}{\textbf{168.0 ± 1.5}}}              &\textcolor{red}{\textbf{0.718 ± 0.005}}               \\ \hline \hline 
		\multirow{6}{*}{\textbf{\rotatebox{90}{Service call}}} & \texttt{MVURE}                                         & \multicolumn{1}{l|}{1428 ± 33}    & \multicolumn{1}{l|}{2180 ± 46}     & 0.367 ± 0.027 & \multicolumn{1}{l|}{190.3 ± 9.8}  & \multicolumn{1}{l|}{266.9 ± 12.1}  & 0.441 ± 0.050 & \multicolumn{1}{l|}{102.1 ± 4.8}  & \multicolumn{1}{l|}{164.7 ± 2.7}   & 0.479 ± 0.017 \\ \cline{2-11} 
		& \texttt{MGFN}                                          & \multicolumn{1}{l|}{1554 ± 81}    & \multicolumn{1}{l|}{2286 ± 115}    & 0.303 ± 0.069 & \multicolumn{1}{l|}{208.2 ± 11.3} & \multicolumn{1}{l|}{293.4 ± 16.6}  & 0.329 ± 0.077 & \multicolumn{1}{l|}{102.8 ± 2.2}  & \multicolumn{1}{l|}{166.3 ± 2.5}   & 0.468 ± 0.021 \\ \cline{2-11} 
		& \texttt{RDCL}                                     & \multicolumn{1}{l|}{1783 ± 21}    & \multicolumn{1}{l|}{2597 ± 38}     & 0.103 ± 0.026 & \multicolumn{1}{l|}{195.7 ± 7.6}  & \multicolumn{1}{l|}{272.1 ± 10.1}  & 0.445 ± 0.041 & \multicolumn{1}{l|}{116.6 ± 2.3}  & \multicolumn{1}{l|}{196.7 ± 3.2}   & 0.256 ± 0.024 \\ \cline{2-11} 
		& \texttt{HREP}                                          & \multicolumn{1}{l|}{1430 ± 29}    & \multicolumn{1}{l|}{2286 ± 34}     & 0.398 ± 0.021 & \multicolumn{1}{l|}{185.7 ± 6.1}  & \multicolumn{1}{l|}{262.2 ± 10.8}  & 0.468 ± 0.022 & \multicolumn{1}{l|}{103.4 ± 3.2}  & \multicolumn{1}{l|}{167.4 ± 4.6}   & 0.461 ± 0.029 \\ \cline{2-11} 
		& \texttt{HAF}                                      & \multicolumn{1}{l|}{\textcolor{blue}{\underline{1273 ± 20}}}    & \multicolumn{1}{l|}{\textcolor{blue}{\underline{1951 ± 27}}}     & \textcolor{blue}{\underline{0.493 ± 0.014}} & \multicolumn{1}{l|}{\textcolor{blue}{\underline{159.3 ± 13.9}}} & \multicolumn{1}{l|}{\textcolor{blue}{\underline{222.0 ± 18.9}}}  & \textcolor{blue}{\underline{0.613 ± 0.067}} & \multicolumn{1}{l|}{\textcolor{blue}{\underline{81.5 ± 2.5}}}   & \multicolumn{1}{l|}{\textcolor{blue}{\underline{142.1 ± 3.2}}}   & \textcolor{blue}{\underline{0.612 ± 0.018}} \\ \cline{2-11} 
		& \texttt{CURE}                                     & \multicolumn{1}{l|}{\textcolor{red}{\textbf{1244 ± 12}}}             & \multicolumn{1}{l|}{\textcolor{red}{\textbf{1913 ± 4}}}              &\textcolor{red}{\textbf{0.513 ± 0.002}}               & \multicolumn{1}{l|}{\textcolor{red}{\textbf{131.0 ± 2.8}}}             & \multicolumn{1}{l|}{\textcolor{red}{\textbf{184.3 ± 2.2}}}              &\textcolor{red}{
			\textbf{0.736 ± 0.006}}               & \multicolumn{1}{l|}{\textcolor{red}{\textbf{73.8 ± 1.6}}}             & \multicolumn{1}{l|}{\textcolor{red}{\textbf{120.3 ± 1.5}}}              &\textcolor{red}{\textbf{0.722 ± 0.007}}               \\ \bottomrule[1.5pt]

	\end{tabular}
	\label{tab:main}
	\vspace{-10pt}
\end{table*}
\noindent
\textbf{Datasets.}
We evaluate \texttt{CURE} on three metropolitan-scale urban datasets collected from New York City (\textbf{NY})\footnote{\url{https://opendata.cityofnewyork.us/}}, Chicago (\textbf{Chi})\footnote{\url{https://data.cityofchicago.org/}}, and San Francisco (\textbf{SF})\footnote{\url{https://datasf.org/opendata/}}. These datasets cover diverse urban environments with different spatial layouts, population distributions, and activity patterns. Following prior work~\cite{DBLP:conf/icde/Sun0CFKT24}, each city is partitioned into spatial regions, which serve as the basic units for feature construction and downstream analytics.

For each city, we construct heterogeneous urban views from multiple data sources. The mobility view is derived from taxi trip records aggregated into region-level origin--destination flows. The POI view is constructed from OpenStreetMap\footnote{\url{https://www.openstreetmap.org/}} and characterizes the functional composition of each region. The land-use view describes planning-oriented regional attributes and reflects relatively stable urban structures. We further collect three types of region-level urban activity records, including Foursquare check-ins\footnote{\url{https://sites.google.com/site/yangdingqi/home/foursquare-dataset}}, crime incidents, and civic service calls, which are used as prediction targets. Dataset statistics are reported in Table~\ref{tab:dataset}.

\noindent
\textbf{Evaluation Scenarios.}
We evaluate the learned region representations on three analytics tasks. \emph{Check-in prediction} estimates the volume of user check-ins in each region and reflects human mobility and location-based activity intensity. \emph{Crime forecasting} predicts regional crime occurrences and evaluates whether the embeddings capture public-safety-related spatial patterns. \emph{Service call prediction} estimates civic service requests and measures the ability to model urban demand and operational needs.

Following prior studies~\cite{DBLP:conf/ijcai/0004LLH20,DBLP:conf/aaai/ZhouHCS023,DBLP:conf/ijcai/WuYFPZZ0W22,DBLP:conf/icde/Sun0CFKT24}, all tasks are formulated as region-level regression problems. The learned region embeddings are used as input features to a lightweight regression model that predicts aggregated activity counts. We use Lasso regression~\cite{DBLP:conf/aisd/KumarSP23} as the downstream predictor, since its sparsity regularization mitigates overfitting and provides a fair evaluation of embedding quality.

\noindent
\textbf{Evaluation Metrics.}
We use three standard regression metrics: Mean Absolute Error (MAE): $\mathrm{MAE}=\frac{1}{n}\sum_{i=1}^{n}|y_i-\hat{y}_i|$ , Root Mean Square Error (RMSE): $\mathrm{RMSE}=\sqrt{\frac{1}{n}\sum_{i=1}^{n}(y_i-\hat{y}_i)^2}$, and the coefficient of determination ($R^2$): $R^{2}=1-\frac{\sum_{i=1}^{n}(y_i-\hat{y}_i)^2}
{\sum_{i=1}^{n}(y_i-\bar{y})^2}$,
where $n$ is the number of regions, $y_i$ and $\hat{y}_i$ are the ground-truth and predicted values of region $\mathcal{R}_i$, and $\bar{y}$ is the mean ground-truth value. Lower MAE and RMSE and higher $R^2$ indicate better performance.

\noindent
\textbf{Baselines.}
We compare \texttt{CURE} with five representative urban region representation methods. \texttt{MVURE}~\cite{DBLP:conf/ijcai/0004LLH20} jointly models mobility patterns and regional attributes for multi-view region embedding. \texttt{MGFN}~\cite{DBLP:conf/ijcai/WuYFPZZ0W22} constructs multiple mobility graphs to capture diverse movement patterns. \texttt{RegionDCL} (\texttt{RDCL})~\cite{DBLP:conf/kdd/LiHCW023} uses dual contrastive learning over building footprints and POI information.  \texttt{HREP}~\cite{DBLP:conf/aaai/ZhouHCS023} models heterogeneous urban relations with relation-aware graph convolution. \texttt{HAFusion} (\texttt{HAF})~\cite{DBLP:conf/icde/Sun0CFKT24} integrates intra-view, inter-view, and cross-region correlations through attentive feature learning and fusion. These baselines cover mobility-based, semantic-based, heterogeneous graph-based, contrastive, and attention-based representation paradigms.

\begin{table*}[t]
	\centering
	\small
	\setlength{\tabcolsep}{1pt} %
	\renewcommand{\arraystretch}{1.9}  %
	\caption{Ablation Studies for \texttt{CURE}.}
	\begin{tabular}{l|l|lll|lll|lll}
		\toprule[1.5pt]
		\multirow{2}{*}{\textbf{}}             & \multicolumn{1}{c|}{\multirow{2}{*}{\textbf{Method}}} & \multicolumn{3}{c|}{\textbf{NY}}                                            & \multicolumn{3}{c|}{\textbf{Chi}}                                                  & \multicolumn{3}{c}{\textbf{SF}}                                            \\ \cline{3-11} 
		& \multicolumn{1}{c|}{}                                  & \multicolumn{1}{c|}{\textbf{MAE} ($\downarrow$)} & \multicolumn{1}{c|}{\textbf{RMSE} ($\downarrow$)} & \multicolumn{1}{c|}{\textbf{$R^2$} ($\uparrow$)}  & \multicolumn{1}{c|}{\textbf{MAE} ($\downarrow$)} & \multicolumn{1}{c|}{\textbf{RMSE} ($\downarrow$)} &\multicolumn{1}{c|}{ \textbf{$R^2$} ($\uparrow$)}   & \multicolumn{1}{c|}{\textbf{MAE} ($\downarrow$)} & \multicolumn{1}{c|}{\textbf{RMSE} ($\downarrow$)} & \multicolumn{1}{c}{\textbf{$R^2$} ($\uparrow$)}   \\ \hline \hline
		\multirow{4}{*}{\textbf{\rotatebox{90}{Check-in}}}     
		& \texttt{w/o GIVE}                                      & \multicolumn{1}{l|}{194.9 ± 3.4}    & \multicolumn{1}{l|}{292.9 ± 5.3}     &0.872 ± 0.005  & \multicolumn{1}{l|}{1000 ± 75.4} & \multicolumn{1}{l|}{1906 ± 64.4}  &0.876 ± 0.008  & \multicolumn{1}{l|}{228.1 ± 8.2}   & \multicolumn{1}{l|}{393.5 ± 18.6}   &0.844 ± 0.015  \\ \cline{2-11} 
		& \texttt{w/o CIVI}                                      & \multicolumn{1}{l|}{195.7 ± 5.9}    & \multicolumn{1}{l|}{311.7 ± 10.6}     &0.855 ± 0.010  & \multicolumn{1}{l|}{998 ± 90.4} & \multicolumn{1}{l|}{1912 ± 150.5}  &0.874 ± 0.019  & \multicolumn{1}{l|}{242.9 ± 8.5}   & \multicolumn{1}{l|}{430.9 ± 21.3}   &0.813 ± 0.018  \\ \cline{2-11} 
		& \texttt{w/o HGRF}                                      & \multicolumn{1}{l|}{209.0 ± 1.9}    & \multicolumn{1}{l|}{327.5 ± 3.4}     &0.839 ± 0.003  & \multicolumn{1}{l|}{1037 ± 68.3} & \multicolumn{1}{l|}{1947 ± 72.4}  &0.870 ± 0.010  & \multicolumn{1}{l|}{227.8 ± 4.8}   & \multicolumn{1}{l|}{421.1 ± 19.5}   &0.821 ± 0.017  \\ \cline{2-11}  
		& \texttt{CURE}                                     & \multicolumn{1}{l|}{\textcolor{red}{\textbf{186.7 ± 2.2}}}             & \multicolumn{1}{l|}{\textcolor{red}{\textbf{289.0 ± 4.6}}}              & \textcolor{red}{\textbf{0.875 ± 0.004}}              & \multicolumn{1}{l|}{\textcolor{red}{\textbf{762 ± 56}}}             & \multicolumn{1}{l|}{\textcolor{red}{\textbf{1569 ± 19}}}              & \textcolor{red}{\textbf{0.916 ± 0.011}}    & \multicolumn{1}{l|}{\textcolor{red}{\textbf{215.8 ± 5.5}}}             & \multicolumn{1}{l|}{\textcolor{red}{\textbf{373.4 ± 6.0}}}              &\textcolor{red}{\textbf{0.860 ± 0.004}}               \\ \hline \hline
		
		\multirow{4}{*}{\textbf{\rotatebox{90}{Crime}}}        
		& \texttt{w/o GIVE}                                      & \multicolumn{1}{l|}{ 60.0 ± 2.0}    & \multicolumn{1}{l|}{81.5 ± 2.4}     &0.696 ± 0.018  & \multicolumn{1}{l|}{ 83.0 ± 4.8} & \multicolumn{1}{l|}{110.6 ± 5.6}  &0.605 ± 0.040  & \multicolumn{1}{l|}{103.1 ± 2.6}   & \multicolumn{1}{l|}{174.9 ± 3.6}   &0.694 ± 0.013  \\ \cline{2-11} 
		& \texttt{w/o CIVI}                                      & \multicolumn{1}{l|}{57.4 ± 1.3}    & \multicolumn{1}{l|}{ 79.9 ± 1.8}     &0.708 ± 0.013
		& \multicolumn{1}{l|}{78.4 ± 4.6} & \multicolumn{1}{l|}{107.5 ± 5.8}  &0.627 ± 0.041  & \multicolumn{1}{l|}{107.5 ± 3.5}   & \multicolumn{1}{l|}{186.6 ± 2.7}   &0.652 ± 0.010  \\ \cline{2-11} 
		& \texttt{w/o HGRF}                                      & \multicolumn{1}{l|}{57.0 ± 0.3}    & \multicolumn{1}{l|}{79.0 ± 1.7}     &0.714 ± 0.012  & \multicolumn{1}{l|}{78.0 ± 1.0} & \multicolumn{1}{l|}{107.8 ± 3.9}  &0.625 ± 0.027 & \multicolumn{1}{l|}{105.6 ± 5.1}   & \multicolumn{1}{l|}{176.4 ± 0.7}   &0.689 ± 0.002  \\ \cline{2-11} 
		& \texttt{CURE}                                     & \multicolumn{1}{l|}{\textcolor{red}{\textbf{53.2 ± 1.5}}}             & \multicolumn{1}{l|}{\textcolor{red}{\textbf{73.1 ± 0.9}}}              &\textcolor{red}{\textbf{0.755 ± 0.006}}               & \multicolumn{1}{l|}{\textcolor{red}{\textbf{69.6 ± 3.2}}}             & \multicolumn{1}{l|}{\textcolor{red}{\textbf{91.2 ± 0.9}}}              &\textcolor{red}{\textbf{0.732 ± 0.005}}               & \multicolumn{1}{l|}{\textcolor{red}{\textbf{99.1 ± 0.6}}}             & \multicolumn{1}{l|}{\textcolor{red}{\textbf{168.0 ± 1.5}}}              &\textcolor{red}{\textbf{0.718 ± 0.005}}               \\ \hline \hline
		
		\multirow{4}{*}{\textbf{\rotatebox{90}{Service call}}} 
		& \texttt{w/o GIVE}                                      & \multicolumn{1}{l|}{1267 ± 9.6}    & \multicolumn{1}{l|}{1920 ± 17.8}     &0.510 ± 0.009  & \multicolumn{1}{l|}{163.7 ± 7.2} & \multicolumn{1}{l|}{226.1 ± 4.1}  &0.603 ± 0.014  & \multicolumn{1}{l|}{82.6 ± 5.0}   & \multicolumn{1}{l|}{137.8 ± 10.0}   &0.633 ± 0.054  \\ \cline{2-11} 
		& \texttt{w/o CIVI}                                      & \multicolumn{1}{l|}{1278 ± 6.4}    & \multicolumn{1}{l|}{1965 ± 4.3}     &0.486 ± 0.002  & \multicolumn{1}{l|}{161.4 ± 2.9} & \multicolumn{1}{l|}{225.7 ± 1.7}  &0.605 ± 0.006  & \multicolumn{1}{l|}{84.0 ± 2.3}   & \multicolumn{1}{l|}{145.0 ± 4.2}   &0.596 ± 0.023  \\ \cline{2-11} 
		& \texttt{w/o HGRF}                                      & \multicolumn{1}{l|}{1347 ± 75.8}    & \multicolumn{1}{l|}{2031 ± 90.1}     &0.450 ± 0.050  & \multicolumn{1}{l|}{153.8 ± 5.9} & \multicolumn{1}{l|}{214.0 ± 5.7}  &0.644 ± 0.019  & \multicolumn{1}{l|}{78.7 ± 2.8}   & \multicolumn{1}{l|}{134.1 ± 7.8}   &0.654 ± 0.041  \\ \cline{2-11} 
		& \texttt{CURE}                                     & \multicolumn{1}{l|}{\textcolor{red}{\textbf{1244 ± 12}}}             & \multicolumn{1}{l|}{\textcolor{red}{\textbf{1913 ± 4}}}              &\textcolor{red}{\textbf{0.513 ± 0.002}}               & \multicolumn{1}{l|}{\textcolor{red}{\textbf{131.0 ± 2.8}}}             & \multicolumn{1}{l|}{\textcolor{red}{\textbf{184.3 ± 2.2}}}              &\textcolor{red}{
			\textbf{0.736 ± 0.006}}               & \multicolumn{1}{l|}{\textcolor{red}{\textbf{73.8 ± 1.6}}}             & \multicolumn{1}{l|}{\textcolor{red}{\textbf{120.3 ± 1.5}}}              &\textcolor{red}{\textbf{0.722 ± 0.007}}               \\ \bottomrule[1.5pt]
	\end{tabular}
	\label{tab:abl}
	\vspace{-10pt}
\end{table*}
\noindent
\textbf{Implementation Details.}
We implement \texttt{CURE} using Python 3.8.18 and PyTorch 1.10, and conduct experiments on an NVIDIA RTX 8000 GPU. The hidden dimension is set to 256. Multi-head attention is used in the graph-guided intra-view encoder, confounder-aware inter-view interaction module, and region-level refinement module. The numbers of intra-view encoding blocks, inter-view interaction blocks, and region fusion blocks are selected using a validation set. Dropout is applied after input projection and within attention-based blocks.

All methods are evaluated under the same data split and downstream regression protocol. Models are trained for at most 2000 epochs with early stopping based on validation performance. We use Adam with an initial learning rate of $5\times10^{-4}$ and weight decay for regularization.

\subsection{Main Results}

\noindent
\textbf{Check-in Prediction.}
We first evaluate all methods on check-in prediction, which measures whether the learned representations capture human mobility and location-activity patterns. As shown in Table~\ref{tab:main}, \texttt{CURE} consistently achieves the best performance across all cities and metrics. Compared with the strongest baseline \texttt{HAF}, \texttt{CURE} reduces MAE from 202.8 to 186.7 and RMSE from 322.8 to 289.0 in NY, while improving $R^2$ from 0.844 to 0.875. In Chi, \texttt{CURE} improves $R^2$ from 0.870 to 0.916 and reduces RMSE from 1947 to 1569. In SF, it increases $R^2$ from 0.813 to 0.860. These results indicate that \texttt{CURE} more effectively captures mobility-related urban semantics and cross-region dependencies than correlation-driven baselines.

\noindent
\textbf{Crime Forecasting.}
We next evaluate crime forecasting, which requires modeling public-safety-related spatial patterns and socio-environmental context. Table~\ref{tab:main} shows that \texttt{CURE} again outperforms all baselines. In NY, \texttt{CURE} improves $R^2$ from 0.734 to 0.755 over \texttt{HAF}, while reducing MAE from 56.1 to 53.2 and RMSE from 76.1 to 73.1. The improvement is more substantial in Chi, where $R^2$ increases from 0.631 to 0.732 and RMSE decreases from 107.1 to 91.2. In SF, \texttt{CURE} achieves the highest $R^2$ of 0.718. These gains suggest that confounder-aware multi-view integration helps capture localized urban risks and spatially dependent social activities.

\noindent
\textbf{Service Call Prediction.}
Finally, we evaluate service call prediction, which reflects civic service demand and operational urban needs. This task is challenging because service requests are often event-driven and spatially irregular. As reported in Table~\ref{tab:main}, \texttt{CURE} achieves the best results on all three datasets. In NY, it reduces MAE from 1273 to 1244 and RMSE from 1951 to 1913 over \texttt{HAF}, while increasing $R^2$ from 0.493 to 0.513. Larger improvements are observed in Chi, where $R^2$ increases from 0.613 to 0.736 and RMSE decreases from 222.0 to 184.3. In SF, \texttt{CURE} improves $R^2$ from 0.612 to 0.722 and reduces RMSE from 142.1 to 120.3. These results demonstrate that \texttt{CURE} is effective for heterogeneous urban analytics tasks with different spatial regularities and uncertainty levels.

\subsection{Evaluation on the Chengdu Dataset}

We additionally evaluate \texttt{CURE} on the Chengdu dataset, which contains 836 regions. The dataset provides three types of regional information, namely mobility outflow, mobility inflow, and POI attributes, which can be naturally regarded as three complementary views under the \texttt{CURE} framework.
For each view, we construct a view-specific regional graph and apply \texttt{CURE}'s shared-component estimation, residualization, and multi-view fusion mechanisms. We evaluate the resulting representations on three downstream urban socioeconomic prediction tasks: carbon emissions, GDP, and population.

As shown in Table~\ref{tab:cd}, \texttt{CURE} consistently achieves competitive performance across all three tasks, demonstrating that the proposed framework remains effective when applied to a substantially larger and heterogeneous multi-view urban dataset.
\begin{table*}[htp]
	\small 
	\centering
	\caption{Performance comparison on the Chengdu (CD) dataset.}
	\label{tab:cd}
	\resizebox{\textwidth}{!}{
		\begin{tabular}{lccccccccc}
			\toprule[1pt]
			& \multicolumn{3}{c}{\textbf{Carbon}}
			& \multicolumn{3}{c}{\textbf{GDP}}
			& \multicolumn{3}{c}{\textbf{Population}} \\
			\cmidrule(lr){2-4}
			\cmidrule(lr){5-7}
			\cmidrule(lr){8-10}
			
			\textbf{Method}
			& \textbf{MAE} & \textbf{RMSE} & $\mathbf{R^2}$
			& \textbf{MAE} & \textbf{RMSE} & $\mathbf{R^2}$
			& \textbf{MAE} & \textbf{RMSE} & $\mathbf{R^2}$ \\
			\midrule
			
			\texttt{HAF}
			& 81.4 & 145.7 & 0.189
			& 198.1 & 350.2 & 0.168
			& 807.6 & 1366.3 & 0.108 \\
			
			\texttt{ComSRE}~\cite{DBLP:conf/kdd/LiJZHC26}
			& 62.3 & 119.5 & 0.412
			& 161.6 & 307.9 & 0.313
			& 600.2 & 1024.5 & 0.432 \\
			
			\texttt{CURE}
			& \textbf{58.2} & \textbf{113.5} & \textbf{0.466}
			& \textbf{158.4} & \textbf{302.5} & \textbf{0.358}
			& \textbf{598.3} & \textbf{1021.8} & \textbf{0.477} \\
			
			\bottomrule[1pt]
		\end{tabular}
	}
\end{table*}

\subsection{Comparison with Shared--Private and Adversarial Multi-View Methods}

Confounder-aware cross-view interaction is a key distinction of \texttt{CURE}. 
Existing urban region representation methods primarily focus on multi-view fusion and cross-view interaction, without explicitly accounting for potentially misleading dependencies induced by latent shared factors. 
To provide a more direct comparison with alternative multi-view representation paradigms, we further consider three representative methods beyond the urban-specific baselines: MISA~\cite{DBLP:conf/mm/HazarikaZP20}, which performs shared--private representation learning; Domain Separation Networks (DSN)~\cite{DBLP:conf/nips/BousmalisTSKE16}, which employs adversarial learning for shared/private feature separation; and MEGAN~\cite{DBLP:conf/ijcai/SunWHTH19}, which adopts adversarial learning for multi-view network representation.

All methods are evaluated on the same datasets, downstream tasks, and evaluation protocol as \texttt{CURE}. 
As shown in Table~\ref{tab:shared_private_baselines}, \texttt{CURE} consistently achieves the best overall performance across New York, Chicago, and San Francisco. 
These results demonstrate that merely separating shared and private representations or employing adversarial multi-view learning is insufficient to address potentially confounding cross-view dependencies. 
In contrast, \texttt{CURE} explicitly attenuates latent shared variation before cross-view interaction while preserving view-specific urban graph structures, leading to more robust regional representations.
\begin{table*}[htp]
	\small 
	\centering
	\caption{Comparison with shared--private and adversarial multi-view representation methods on New York (NY), Chicago (Chi), and San Francisco (SF).}
	\label{tab:shared_private_baselines}
	\resizebox{\textwidth}{!}{
		\begin{tabular}{lccccccccc}
			\toprule
			& \multicolumn{3}{c}{\textbf{NY}}
			& \multicolumn{3}{c}{\textbf{Chi}}
			& \multicolumn{3}{c}{\textbf{SF}} \\
			\cmidrule(lr){2-4}
			\cmidrule(lr){5-7}
			\cmidrule(lr){8-10}
			
			\textbf{Method}
			& \textbf{MAE} & \textbf{RMSE} & $\mathbf{R^2}$
			& \textbf{MAE} & \textbf{RMSE} & $\mathbf{R^2}$
			& \textbf{MAE} & \textbf{RMSE} & $\mathbf{R^2}$ \\
			\midrule
			
			\texttt{MISA}
			& 196.9 & 317.5 & 0.849
			& 854 & 1722 & 0.899
			& 244.5 & 443.8 & 0.802 \\
			
			\texttt{DSN}
			& 201.7 & 310.5 & 0.856
			& 918 & 1960 & 0.869
			& 233.7 & 417.8 & 0.824 \\
			
			\texttt{MEGAN}
			& 209.2 & 327.3 & 0.840
			& 975 & 1931 & 0.873
			& 261.0 & 416.2 & 0.825 \\
			
			\texttt{CURE}
			& \textbf{186.7} & \textbf{289.0} & \textbf{0.875}
			& \textbf{762} & \textbf{1569} & \textbf{0.916}
			& \textbf{215.8} & \textbf{373.4} & \textbf{0.860} \\
			
			\bottomrule
		\end{tabular}
	}
\end{table*}

\subsection{Ablation Study}

To assess the contribution of each component in \texttt{CURE}, we compare the full model with three ablated variants: \texttt{w/o GIVE} removes graph-guided intra-view encoding (\texttt{GIVE}); \texttt{w/o CIVI} removes confounder-aware inter-view interaction (\texttt{CIVI}); and \texttt{w/o HGRF} removes hierarchical graph-aware residual fusion (\texttt{HGRF}). As shown in Table~\ref{tab:abl}, the full \texttt{CURE} consistently outperforms all variants across tasks, cities, and metrics, confirming the effectiveness of the proposed design.

\noindent
\textbf{Effect of GIVE.}
Removing \texttt{GIVE} leads to consistent performance degradation, especially in Chi. For example, on check-in prediction, MAE increases from 762 to 1000 and $R^2$ drops from 0.916 to 0.876. Similar drops are observed in crime forecasting and service call prediction, where Chi $R^2$ decreases from 0.732 to 0.605 and from 0.736 to 0.603, respectively. The results indicate that graph-guided intra-view encoding is essential for preserving view-specific spatial structures and local urban dependencies.

\noindent
\textbf{Effect of CIVI.}
The variant \texttt{w/o CIVI} also underperforms the full model on all tasks. In check-in prediction, $R^2$ decreases from 0.875 to 0.855 in NY, from 0.916 to 0.874 in Chi, and from 0.860 to 0.813 in SF. The degradation is also evident in service call prediction, where SF $R^2$ drops from 0.722 to 0.596. These results demonstrate that confounder-aware inter-view interaction improves the reliability of multi-view integration by mitigating spurious cross-view correlations.

\noindent
\textbf{Effect of HGRF.}
Removing \texttt{HGRF} causes notable declines, particularly on check-in prediction and service call prediction. For check-in prediction, NY MAE increases from 186.7 to 209.0, while Chi MAE increases from 762 to 1037. For service call prediction in NY, $R^2$ decreases from 0.513 to 0.450. These results verify that hierarchical graph-aware residual fusion is important for integrating multi-level urban semantics while retaining useful low-level structural information.

\textbf{Comparison with Alternative Cross-View Interaction Designs: }To further justify the proposed cross-view interaction design, we compare \texttt{CURE} with two simpler alternatives. 
Direct Concatenation (DC) directly concatenates the representations learned from different views without explicit cross-view interaction, while Cross-Attention (CA) replaces the proposed confounder-aware interaction mechanism with standard cross-view attention. 

As shown in Table~\ref{tab:interaction_design}, \texttt{CURE} consistently outperforms both DC and CA across New York, Chicago, and San Francisco. 
These results suggest that the performance gain cannot be attributed solely to increased fusion or interaction capacity. 
Instead, explicitly attenuating shared latent variation before cross-view interaction leads to more effective and robust multi-view regional representations.

\begin{table*}[t]
	\centering
	\caption{Comparison with alternative cross-view interaction designs on New York (NY), Chicago (Chi), and San Francisco (SF).}
	\label{tab:interaction_design}
	\resizebox{\textwidth}{!}{
		\begin{tabular}{lccccccccc}
			\toprule
			& \multicolumn{3}{c}{\textbf{NY}}
			& \multicolumn{3}{c}{\textbf{Chi}}
			& \multicolumn{3}{c}{\textbf{SF}} \\
			\cmidrule(lr){2-4}
			\cmidrule(lr){5-7}
			\cmidrule(lr){8-10}
			
			\textbf{Method}
			& \textbf{MAE} & \textbf{RMSE} & $\mathbf{R^2}$
			& \textbf{MAE} & \textbf{RMSE} & $\mathbf{R^2}$
			& \textbf{MAE} & \textbf{RMSE} & $\mathbf{R^2}$ \\
			\midrule
			
			DC
			& 206.4 & 319.2 & 0.848
			& 897 & 1749 & 0.896
			& 265.1 & 438.3 & 0.806 \\
			
			CA
			& 194.7 & 317.3 & 0.850
			& 795 & 1643 & 0.907
			& 240.6 & 434.1 & 0.810 \\
			
			\textbf{CURE}
			& \textbf{186.7} & \textbf{289.0} & \textbf{0.875}
			& \textbf{762} & \textbf{1569} & \textbf{0.916}
			& \textbf{215.8} & \textbf{373.4} & \textbf{0.860} \\
			
			\bottomrule
		\end{tabular}
	}
\end{table*}
\subsection{Parameter Sensitivity Analysis}

\begin{figure}[t]
	\centering
	\includegraphics[scale=0.18]{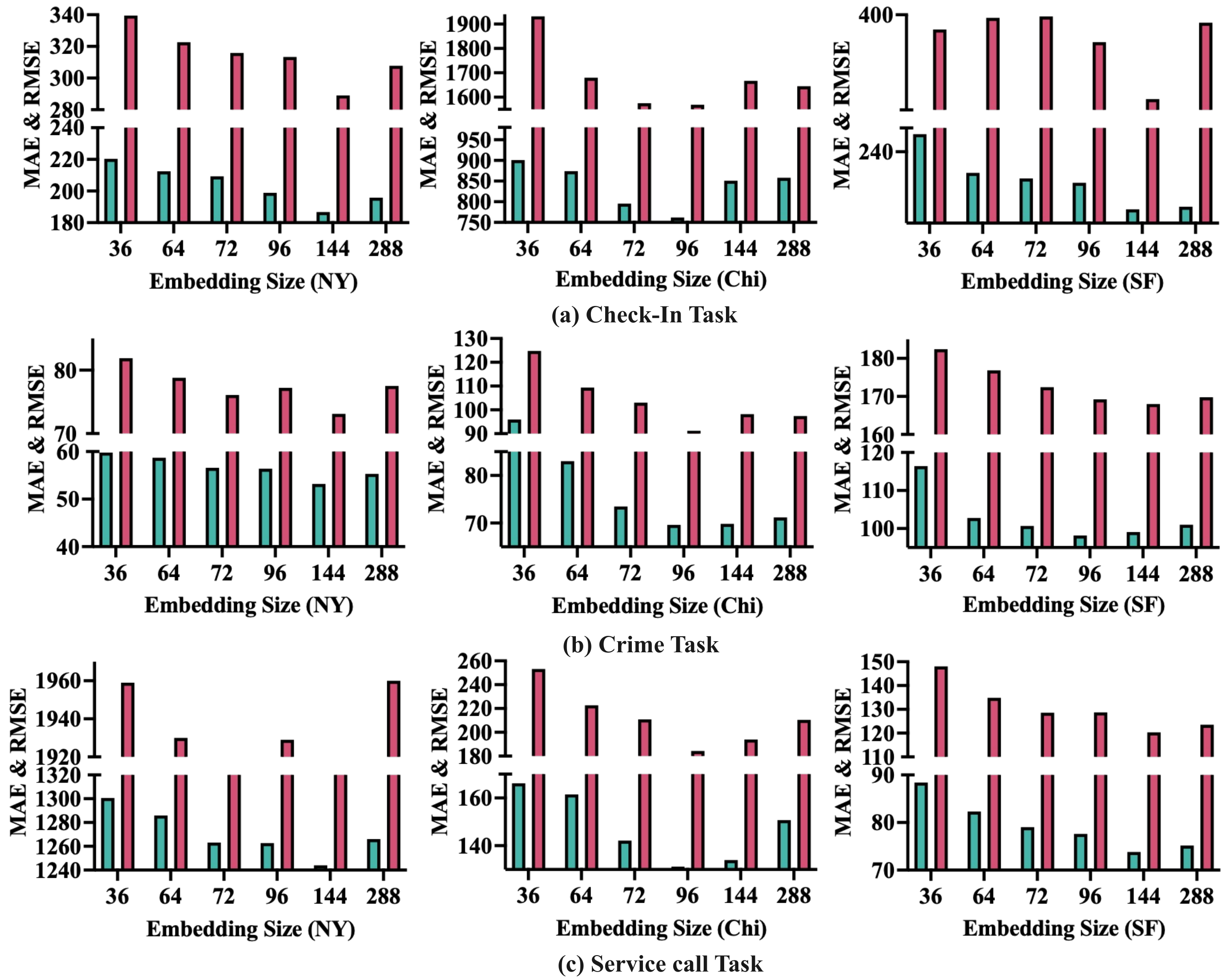}
	\caption{Effects of Embedding Size.}
	\label{fig:ES}
\end{figure}
\noindent
\textbf{Effects of Region Embedding Size.}
We vary the embedding size $d$ in $\{36,72,96,144,288\}$ to examine the sensitivity of \texttt{CURE} to representation dimensionality. As shown in Fig.~\ref{fig:ES}, small embeddings generally yield larger MAE and RMSE, indicating insufficient capacity to encode heterogeneous urban semantics and spatial dependencies. Increasing $d$ improves performance in most cases, but the gains saturate at moderate dimensionalities. Across tasks and cities, $d=96$ and $d=144$ usually achieve the best or near-best results, whereas $d=288$ does not consistently provide further improvement. This suggests that overly large embeddings may introduce redundant dimensions and increase overfitting risk. We therefore select a moderate embedding size in the main experiments to balance expressiveness and generalization.

\begin{figure}[t]
	\centering
	\includegraphics[scale=0.19]{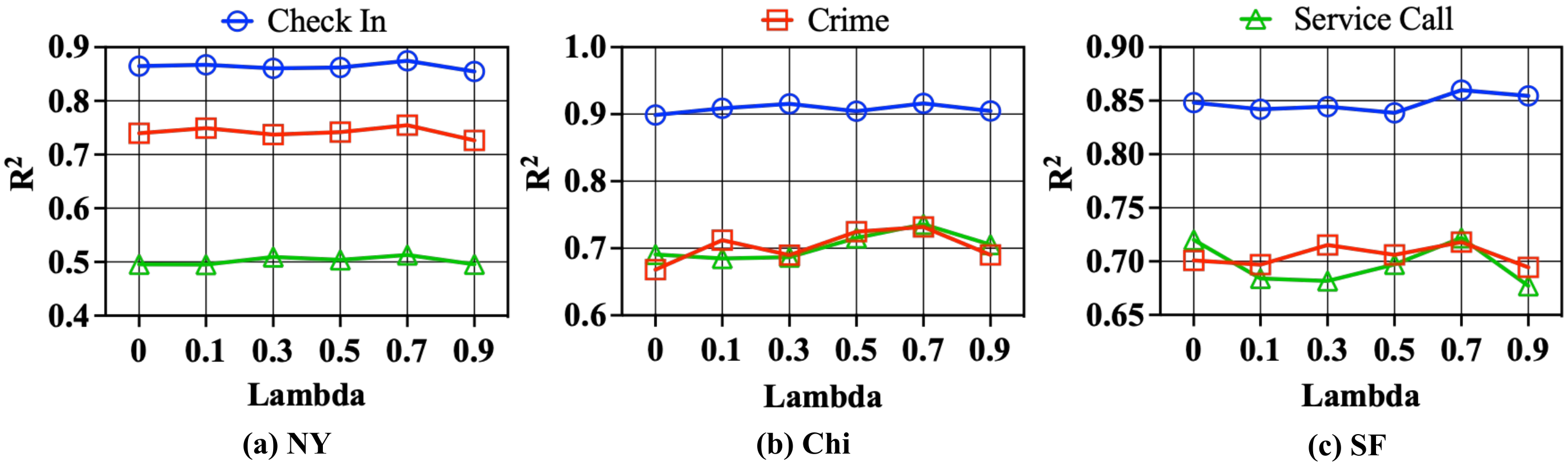}
	\caption{Effects of Lambda ($\lambda$).}
	\label{fig:lambda}
	\vspace{-10pt}
\end{figure}
\noindent
\textbf{Effects of $\lambda$.}
Fig.~\ref{fig:lambda} shows the effect of the deconfounding weight $\lambda$ on the three datasets. In general, the model is stable across different values of $\lambda$, indicating low sensitivity to this hyperparameter. A moderate value of $\lambda$ generally improves performance, especially for crime and service call prediction, suggesting that deconfounding helps reduce biased correlations and learn more robust representations.
Across NY, Chi, and SF, Check-in consistently achieves the highest and most stable $R^2$. For crime and service call, performance usually increases when $\lambda$ grows from 0 to 0.5 or 0.7, but drops when $\lambda$ becomes too large, e.g., when $\lambda=0.9$. This indicates that weak deconfounding may be insufficient, while overly strong deconfounding may remove useful task-related information.
Taken together, $\lambda=0.7$ achieves the best or near-best performance in most cases, providing a good balance between prediction accuracy and confounder-reduced representation learning. Therefore, we set $\lambda=0.7$ as the default value in our experiments.

\subsection{Model Scalability}
\begin{figure}[h]
	\centering
	\includegraphics[scale=0.19]{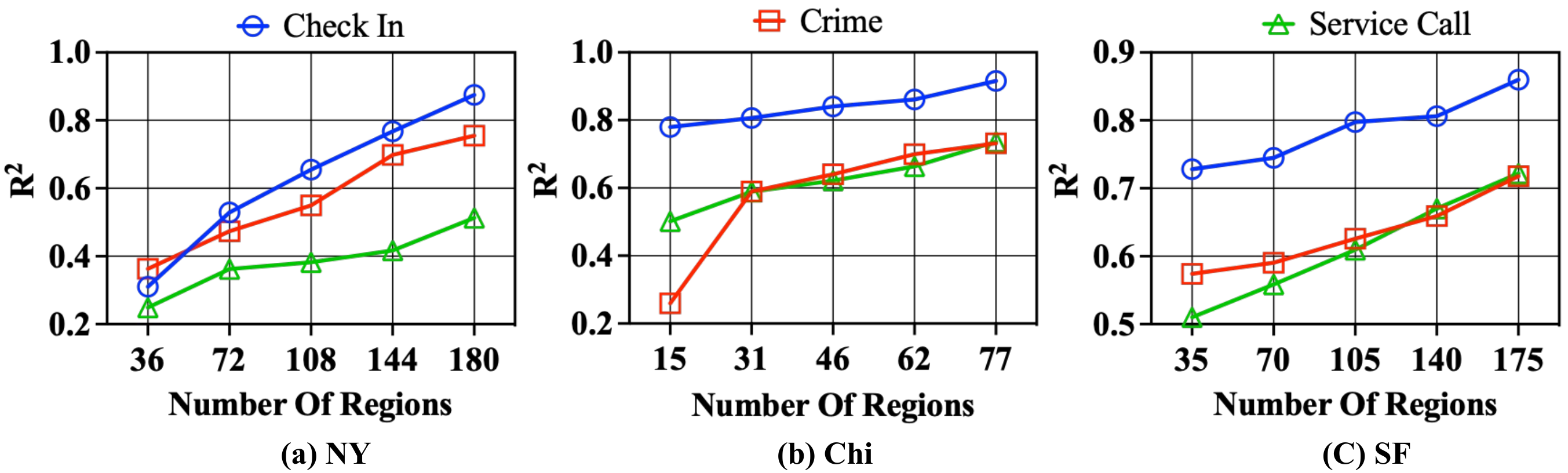}
	\caption{Scalability Analysis with respect to the Number of Regions.}
	\label{fig:NR}
\end{figure}

\noindent
\textbf{Scalability w.r.t. Number of Regions.}
We evaluate the scalability of \texttt{CURE} by varying the number of urban regions and reporting the downstream prediction performance. As shown in Fig.~\ref{fig:NR}, the performance generally improves as the number of regions increases across cities and tasks. This trend indicates that using more regions provides richer spatial coverage and more complete inter-region dependencies, enabling the model to learn more informative urban representations.
The improvement is particularly evident in NY, where the $R^2$ of check-in prediction and crime forecasting increases substantially as the number of regions grows from 36 to 180. Similar trends are observed in SF, where all three tasks benefit from larger region sets. In Chi check-in prediction already achieves relatively high performance with fewer regions, while crime forecasting and service call prediction continue to improve as more regions are included. 

\noindent
\textbf{Computational Scalability.}
Table~\ref{tab:cs} reports the computational cost of \texttt{CURE} on the check-in prediction task under different scale ratios. The scale ratio denotes the proportion of regions sampled from the original city-level region set. AET and PA denote the average training epoch time and peak GPU memory allocation, respectively.
As the scale ratio increases from 0.2 to 1.0, both AET and PA increase consistently across the three datasets. This is expected because larger region sets lead to larger regional graphs and more pairwise region dependencies to be processed during graph-guided encoding and multi-view fusion. Nevertheless, the increase in training time remains moderate. For example, in NY, AET increases from 0.1269s to 0.1590s, while in SF it increases from 0.1153s to 0.1505s. This indicates that \texttt{CURE} can handle larger urban graphs without introducing prohibitive training overhead.
The memory consumption shows a clearer growth pattern than training time. In NY, PA increases from 168.14MB to 328.91MB, and in SF it increases from 167.12MB to 319.89MB as the full region set is used. This trend is mainly due to the storage and propagation of multiple view-specific regional graphs, including mobility, POI, and land-use relations. Since graph-based operations depend on the number of regions and regional connections, larger spatial scales naturally require more GPU memory.
Compared with NY and SF, Chi exhibits lower memory usage and slower growth in AET. This is because Chi contains fewer regions under the same scale ratios, leading to smaller graph structures and lower computation cost. Even at the full scale, Chi only requires 0.0839s per epoch and 117.40MB peak memory allocation. 

\begin{table}[]
	\scriptsize
	\renewcommand{\arraystretch}{1.25} 
	\centering
	\caption{Computational scalability of \texttt{CURE} on the check-in prediction task under different scale ratios. AET denotes the average epoch time, and PA denotes peak GPU memory allocation.}
	\begin{tabular}{l|ll|ll|ll}
		\toprule[1.5pt]
		\multirow{2}{*}{} & \multicolumn{2}{c|}{NY}                                     & \multicolumn{2}{c|}{Chi}                                    & \multicolumn{2}{c}{SF}                                     \\ \cline{2-7} 
		& \multicolumn{1}{c|}{AET (s)} & \multicolumn{1}{c|}{PA (MB)} & \multicolumn{1}{c|}{AET (s)} & \multicolumn{1}{c|}{PA (MB)} & \multicolumn{1}{c|}{AET (s)} & \multicolumn{1}{c}{PA (MB)} \\ \hline
		0.2               & \multicolumn{1}{l|}{0.1269}  & 168.14                       & \multicolumn{1}{l|}{0.0763}  & 88.54                        & \multicolumn{1}{l|}{0.1153}  & 167.12                       \\ \hline
		0.4               & \multicolumn{1}{l|}{0.1315}  & 193.75                       & \multicolumn{1}{l|}{0.0798}  & 92.98                        & \multicolumn{1}{l|}{0.1203}  & 191.55                       \\ \hline
		0.6               & \multicolumn{1}{l|}{0.1393}  & 229.16                       & \multicolumn{1}{l|}{0.0809}  & 100.03                       & \multicolumn{1}{l|}{0.1308}  & 224.47                       \\ \hline
		0.8               & \multicolumn{1}{l|}{0.1546}  & 277.14                       & \multicolumn{1}{l|}{0.0825}  & 107.45                       & \multicolumn{1}{l|}{0.1417}  & 266.38                       \\ \hline
		1.0               & \multicolumn{1}{l|}{0.1590}  & 328.91                       & \multicolumn{1}{l|}{0.0839}  & 117.40                       & \multicolumn{1}{l|}{0.1505}  & 319.89                       \\ \bottomrule[1.5pt]
	\end{tabular}
	\label{tab:cs}
	\vspace{-10pt}
\end{table}

\subsection{Robustness and Reliability Analysis}

We evaluate \texttt{CURE} from two complementary perspectives: \emph{robustness}, which tests whether the learned representations remain stable under missing views and noisy input features, and \emph{reliability}, which examines whether cross-view integration suppresses spurious shared dependencies while preserving stable and task-relevant view-specific information. 

\noindent
\textbf{Robustness to Missing Views.}
\begin{figure}[t]
	\centering
	\includegraphics[scale=0.24]{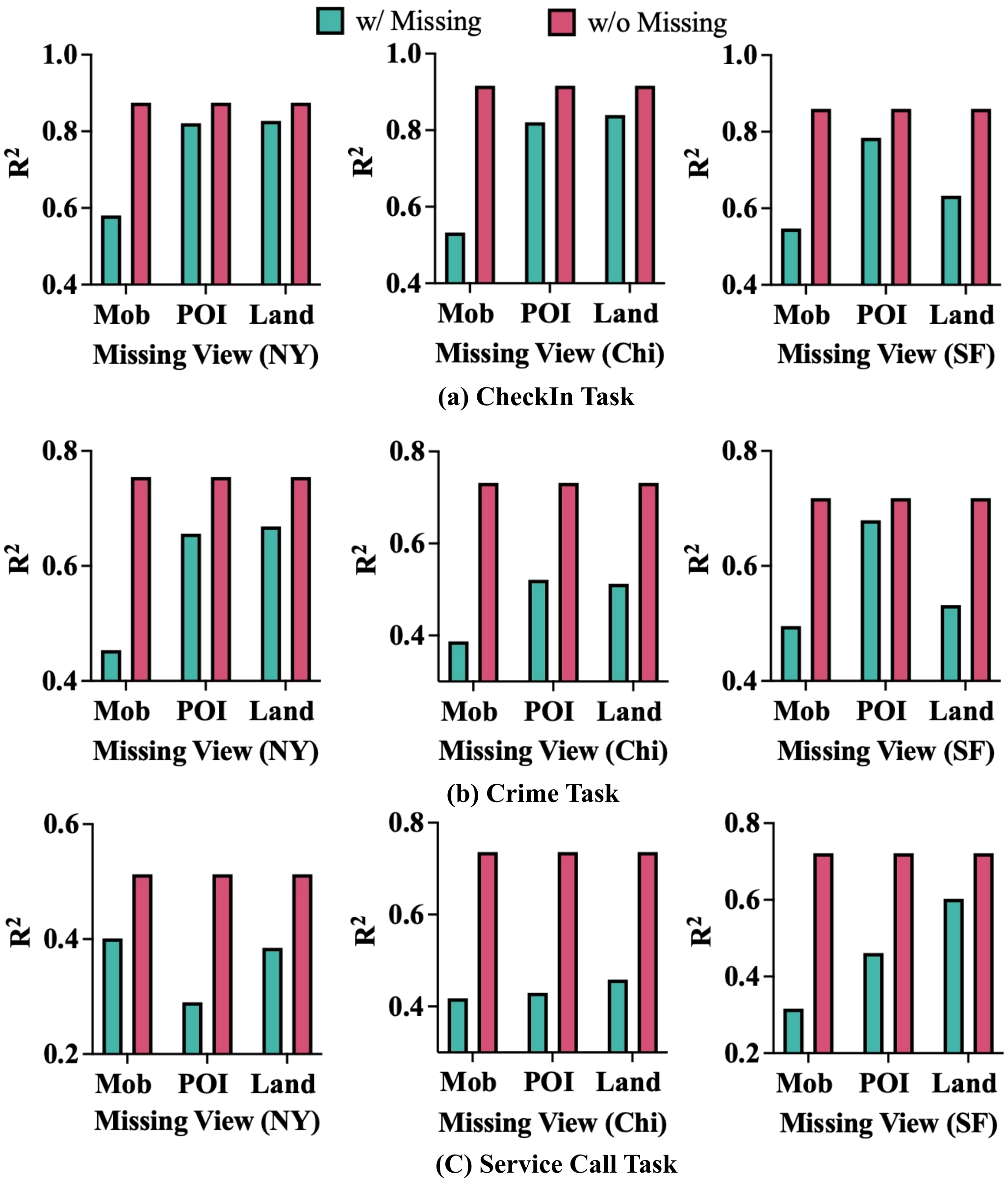}
	\caption{Effects of missing views.}
	\label{fig:MV}
	\vspace{-10pt}
\end{figure}
To evaluate robustness under incomplete urban observations, we mask one input view at a time during inference, including mobility, POI, and land-use views. As shown in Fig.~\ref{fig:MV}, removing any view generally degrades performance, indicating that all three views contribute useful information to region representation learning. The performance drop is often most pronounced when the mobility view is missing, especially for check-in prediction and service call prediction. This is expected because mobility flows directly capture human movement intensity and inter-region interactions, which are closely related to location activity and urban service demand.

In contrast, missing POI or land-use information usually causes a smaller degradation. This suggests that POI and land-use views provide complementary functional and planning-oriented semantics, while mobility supplies more direct dynamic signals for downstream prediction. Nevertheless, \texttt{CURE} maintains reasonable performance when a single view is removed, showing that it can exploit the remaining views rather than depending entirely on one data source. 

\begin{figure}[t]
	\centering
	\includegraphics[scale=0.19]{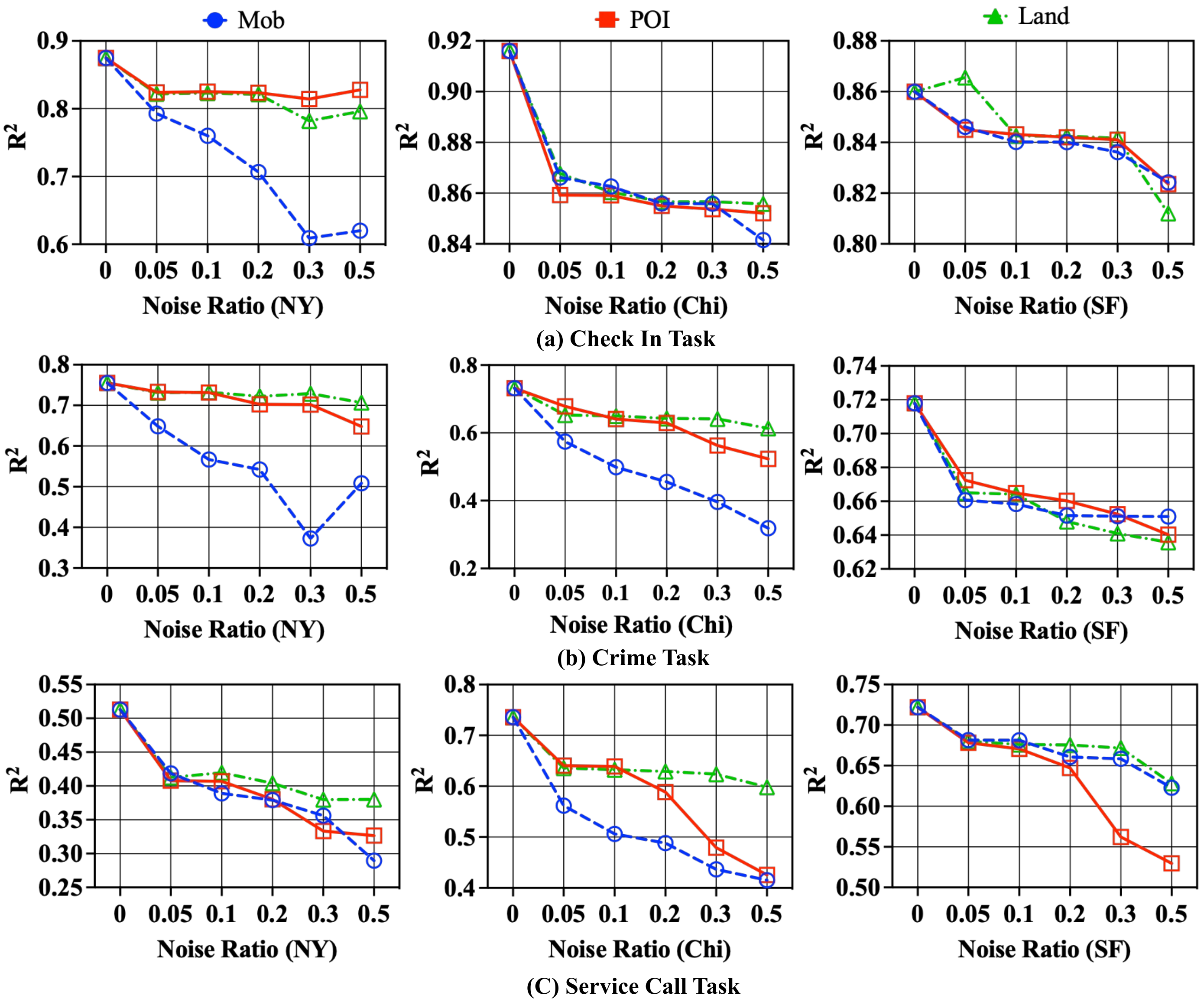}
	\caption{Effects of feature noise.}
	\label{fig:FN}
	\vspace{-10pt}
\end{figure}

\noindent
\textbf{Robustness to Feature Noise.}
We next examine robustness to noisy urban observations by injecting Gaussian noise into each view separately. The noise ratio is varied from $0.05$ to $0.5$, where larger values indicate stronger perturbations. Fig.~\ref{fig:FN} shows that the performance generally decreases as the noise ratio increases, confirming that noisy urban observations can weaken region representation quality. However, the degradation patterns differ across views, tasks, and cities. Perturbing the mobility view often leads to a sharper decline, particularly in tasks that are strongly associated with human movement and spatial interaction. POI and land-use perturbations tend to cause more moderate degradation in several cases, suggesting that semantic and planning-oriented views provide relatively stable contextual information.

Overall, \texttt{CURE} exhibits gradual rather than abrupt performance degradation under feature noise. This indicates that its graph-guided intra-view encoding and hierarchical residual fusion help preserve reliable signals from less corrupted views, while the confounder-aware interaction module reduces over-reliance on unstable cross-view correlations. Together with the missing-view results, the feature-noise study verifies that \texttt{CURE} can learn robust urban region representations under incomplete and noisy multi-view data conditions.

\begin{table}[t]
	\small
	\centering
	\caption{Shared component analysis of \texttt{CURE}. }
	\label{tab:shared_component}
	\setlength{\tabcolsep}{4pt}
	\renewcommand{\arraystretch}{0.95}
	\begin{tabular}{l|l|cc}
		\toprule[1.5pt]
		City & View & SR & RD \\
		\midrule
		\multirow{3}{*}{NY} 
		& POI      & $0.2267 \pm 0.0953$ & $1.51{\times}10^{-8} \pm 5.25{\times}10^{-9}$ \\
		& Land-use & $0.2126 \pm 0.0727$ & $1.41{\times}10^{-8} \pm 4.41{\times}10^{-9}$ \\
		& Mobility & $0.1757 \pm 0.0494$ & $1.21{\times}10^{-8} \pm 2.54{\times}10^{-9}$ \\
		\midrule
		\multirow{3}{*}{Chi} 
		& POI      & $0.1685 \pm 0.0677$ & $1.22{\times}10^{-8} \pm 3.52{\times}10^{-9}$ \\
		& Land-use & $0.2723 \pm 0.1497$ & $1.81{\times}10^{-8} \pm 9.88{\times}10^{-9}$ \\
		& Mobility & $0.2311 \pm 0.0987$ & $1.48{\times}10^{-8} \pm 5.23{\times}10^{-9}$ \\
		\midrule
		\multirow{3}{*}{SF} 
		& POI      & $0.1780 \pm 0.0719$ & $1.26{\times}10^{-8} \pm 3.86{\times}10^{-9}$ \\
		& Land-use & $0.1907 \pm 0.0523$ & $1.30{\times}10^{-8} \pm 2.88{\times}10^{-9}$ \\
		& Mobility & $0.1688 \pm 0.0779$ & $1.18{\times}10^{-8} \pm 3.73{\times}10^{-9}$ \\
		\bottomrule[1.5pt]
	\end{tabular}
	\vspace{-10pt}
\end{table}
\noindent
\textbf{Reliability of Shared Component Separation.}
We further examine whether the deconfounding module reduces shared-factor-induced dependencies, which is a key aspect of reliability defined in the Introduction. Since the true latent factors are unobserved, we use two diagnostic metrics. For each view representation $H^{(v)}$ and its residual representation $R^{(v)}$, the shared component ratio (SR), defined as $\|H^{(v)}-R^{(v)}\|_2/\|H^{(v)}\|_2$, measures how much information is removed as the estimated shared component. The residual dependence (RD), measured by the mean absolute cosine similarity between $R^{(v)}$ and $C$, measures whether the residual representation remains aligned with the shared component. A non-trivial SR together with a small RD indicates that \texttt{CURE} separates shared signals and reduces their influence on residual cross-view interaction. Since representation learning is city-level and task-agnostic, we report these statistics once for each city.

As shown in Table~\ref{tab:shared_component}, all views contain non-negligible shared components, indicating that heterogeneous urban views are not independent but are jointly influenced by latent urban factors. The SR values vary across cities and views. In NY, POI has the largest SR, suggesting that functional urban semantics are more strongly aligned with the estimated shared factor. In Chi, land-use and mobility show higher SR values, indicating stronger coupling between static urban structure, movement patterns, and shared city-level factors. In SF, the SR values are relatively balanced across views, suggesting more evenly distributed shared information among heterogeneous urban signals.
The RD values are consistently close to zero across all cities and views. This confirms that the residualization step effectively removes the component aligned with the estimated shared factor, leaving residual representations that are nearly independent of the shared component.

\noindent
\textbf{Adaptive View Weighting Analysis.}
We further analyze the adaptive fusion weights learned by the hierarchical graph-aware residual fusion module to examine context-dependent view contribution. These weights reflect how \texttt{CURE} allocates importance among confounder-reduced urban views when forming the final region representation. In the context of our reliability definition, such adaptive weighting complements shared component separation by showing whether the model further adjusts residual view contributions according to city- and task-specific contexts, rather than relying on fixed cross-view integration.

\begin{figure}[t]
	\centering
	\includegraphics[scale=0.19]{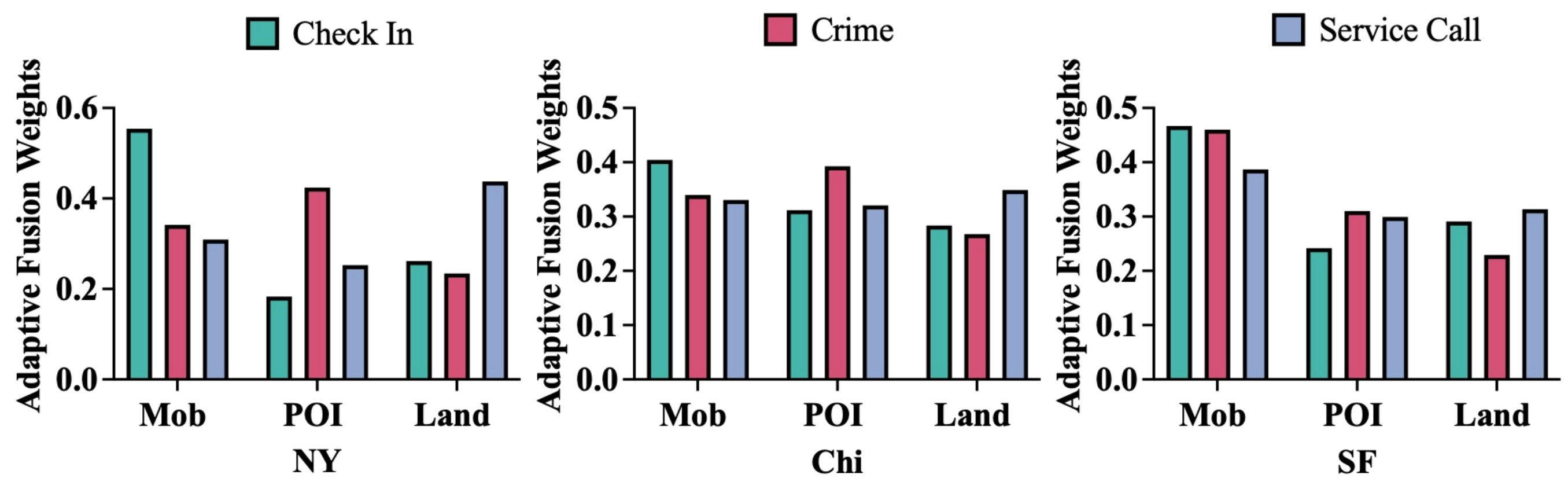}
	\caption{Adaptive Fusion Weights Analysis.}
	\label{fig:VW}
	\vspace{-10pt}
\end{figure}

As shown in Fig.~\ref{fig:VW}, the learned weights vary across both cities and downstream tasks. In NY, check-in prediction assigns the largest weight to mobility, which is consistent with the close relation between check-in activity and human movement. Crime forecasting places more emphasis on POI, suggesting that regional functional composition provides useful signals for public-safety-related prediction. Service call prediction assigns the largest weight to land-use, reflecting the relevance of stable urban structure and planning-oriented attributes. Similar task- and city-dependent patterns are observed in Chi and SF. These results indicate that \texttt{CURE} adaptively integrates residual view representations according to city and task contexts, supporting its ability to model context-dependent view contribution in reliable multi-view data integration.

\section{Conclusion}

We studied reliable multi-view urban data integration for urban region representation learning and proposed \texttt{CURE}, a confounder-aware framework for heterogeneous urban views. By combining graph-guided intra-view encoding, confounder-aware inter-view interaction, and hierarchical graph-aware residual fusion, \texttt{CURE} preserves view-specific structures, reduces shared-factor-induced dependencies, and adaptively integrates stable and task-relevant residual view representations. Experiments on three cities and three tasks show that \texttt{CURE} improves predictive performance, remains robust to incomplete and noisy observations, and supports reliable integration through shared component separation and adaptive view weighting.

\balance
\bibliographystyle{IEEEtran}
\bibliography{IEEEexample}

@inproceedings{DBLP:conf/kdd/SunCTK025,
	author       = {Fengze Sun and
	Yanchuan Chang and
	Egemen Tanin and
	Shanika Karunasekera and
	Jianzhong Qi},
	title        = {FlexiReg: Flexible Urban Region Representation Learning},
	booktitle    = {{KDD}},
	pages        = {2702--2713},
	year         = {2025}
}

@inproceedings{DBLP:conf/kdd/LiHCW023,
	author       = {Yi Li and
	Weiming Huang and
	Gao Cong and
	Hao Wang and
	Zheng Wang},
	title        = {Urban Region Representation Learning with OpenStreetMap Building Footprints},
	booktitle    = { {KDD} },
	pages        = {1363--1373},
	year         = {2023}
}

@inproceedings{DBLP:conf/ijcai/XuZ24,
	author       = {Zhuo Xu and
	Xiao Zhou},
	title        = {{CGAP:} Urban Region Representation Learning with Coarsened Graph
	Attention Pooling},
	booktitle    = {{IJCAI} },
	pages        = {7518--7526},
	year         = {2024}
}

@inproceedings{DBLP:conf/ijcai/0004LLH20,
	author       = {Mingyang Zhang and
	Tong Li and
	Yong Li and
	Pan Hui},
	title        = {Multi-View Joint Graph Representation Learning for Urban Region Embedding},
	booktitle    = { {IJCAI}},
	pages        = {4431--4437},
	year         = {2020}
}

@inproceedings{DBLP:conf/icde/Sun0CFKT24,
	author       = {Fengze Sun and
	Jianzhong Qi and
	Yanchuan Chang and
	Xiaoliang Fan and
	Shanika Karunasekera and
	Egemen Tanin},
	title        = {Urban Region Representation Learning with Attentive Fusion},
	booktitle    = { {ICDE} },
	pages        = {4409--4421},
	year         = {2024}
}

@article{DBLP:journals/tkde/GongWGLLZWZL24,
	author       = {Letian Gong and
	Huaiyu Wan and
	Shengnan Guo and
	Xiucheng Li and
	Yan Lin and
	Erwen Zheng and
	Tianyi Wang and
	Zeyu Zhou and
	Youfang Lin},
	title        = {Spatial-Temporal Cross-View Contrastive Pre-Training for Check-in
	Sequence Representation Learning},
	journal      = {{IEEE} Trans. Knowl. Data Eng.},
	volume       = {36},
	number       = {12},
	pages        = {9308--9321},
	year         = {2024}
}

@article{DBLP:journals/tkde/GongGLLZSLHW25,
	author       = {Letian Gong and
	Shengnan Guo and
	Yan Lin and
	Yichen Liu and
	Erwen Zheng and
	Yiwei Shuang and
	Youfang Lin and
	Jilin Hu and
	Huaiyu Wan},
	title        = {{STCDM:} Spatio-Temporal Contrastive Diffusion Model for Check-In
	Sequence Generation},
	journal      = {{IEEE} Trans. Knowl. Data Eng.},
	volume       = {37},
	number       = {4},
	pages        = {2141--2154},
	year         = {2025}
}

@article{DBLP:journals/tcss/ButtLAS25,
	author       = {Umair Muneer Butt and
	Sukumar Letchmunan and
	Mubashir Ali and
	Hafiz Husnain Raza Sherazi},
	title        = {{START:} {A} Spatiotemporal Autoregressive Transformer for Enhancing
	Crime Prediction Accuracy},
	journal      = {{IEEE} Trans. Comput. Soc. Syst.},
	volume       = {12},
	number       = {6},
	pages        = {4650--4664},
	year         = {2025}
}

@inproceedings{DBLP:conf/aaai/WangLYSYS22,
	author       = {Chenyu Wang and
	Zongyu Lin and
	Xiaochen Yang and
	Jiao Sun and
	Mingxuan Yue and
	Cyrus Shahabi},
	title        = {{HAGEN:} Homophily-Aware Graph Convolutional Recurrent Network for
	Crime Forecasting},
	booktitle    = { {AAAI} },
	pages        = {4193--4200},
	year         = {2022}
}

@inproceedings{DBLP:conf/icde/LiHXXP22,
	author       = {Zhonghang Li and
	Chao Huang and
	Lianghao Xia and
	Yong Xu and
	Jian Pei},
	title        = {Spatial-Temporal Hypergraph Self-Supervised Learning for Crime Prediction},
	booktitle    = { {ICDE}},
	pages        = {2984--2996},
	year         = {2022}
}

@inproceedings{DBLP:conf/aaai/ZhaoFLT22,
	author       = {Xiangyu Zhao and
	Wenqi Fan and
	Hui Liu and
	Jiliang Tang},
	title        = {Multi-Type Urban Crime Prediction},
	booktitle    = {{AAAI} },
	pages        = {4388--4396},
	publisher    = {{AAAI} Press},
	year         = {2022}
}

@article{DBLP:journals/tkde/ZhaoLCZ23,
	author       = {Shuai Zhao and
	Ruiqiang Liu and
	Bo Cheng and
	Daxing Zhao},
	title        = {Classification-Labeled Continuousization and Multi-Domain Spatio-Temporal
	Fusion for Fine-Grained Urban Crime Prediction},
	journal      = {{IEEE} Trans. Knowl. Data Eng.},
	volume       = {35},
	number       = {7},
	pages        = {6725--6738},
	year         = {2023}
}

@article{DBLP:journals/corr/abs-2511-20729,
	author       = {Sean Bin Yang and
	Ying Sun and
	Yunyao Cheng and
	Yan Lin and
	Kristian Torp and
	Jilin Hu},
	title        = {Spatio-Temporal Trajectory Foundation Model - Recent Advances and
	Future Directions},
	journal      = {CoRR},
	volume       = {abs/2511.20729},
	year         = {2025}
}

@article{DBLP:journals/corr/abs-2106-09373,
	author       = {Sean Bin Yang and
	Chenjuan Guo and
	Jilin Hu and
	Jian Tang and
	Bin Yang},
	title        = {Unsupervised Path Representation Learning with Curriculum Negative
	Sampling},
	journal      = {CoRR},
	volume       = {abs/2106.09373},
	year         = {2021}
}

@inproceedings{DBLP:conf/aaai/ZhouHCS023,
	author       = {Silin Zhou and
	Dan He and
	Lisi Chen and
	Shuo Shang and
	Peng Han},
	title        = {Heterogeneous Region Embedding with Prompt Learning},
	booktitle    = {{AAAI} },
	pages        = {4981--4989},
	year         = {2023}
}

@inproceedings{DBLP:conf/ijcai/WuYFPZZ0W22,
	author       = {Shangbin Wu and
	Xu Yan and
	Xiaoliang Fan and
	Shirui Pan and
	Shichao Zhu and
	Chuanpan Zheng and
	Ming Cheng and
	Cheng Wang},
	title        = {Multi-Graph Fusion Networks for Urban Region Embedding},
	booktitle    = {{IJCAI}},
	pages        = {2312--2318},
	year         = {2022}
}

@inproceedings{DBLP:conf/aisd/KumarSP23,
	author       = {Sanjay Kumar and
	Meenakshi Srivastava and
	Vijay Prakash},
	title        = {Comparative Analysis of ARIMA, Deep Learning, and Lasso Regression
	Models for Time Series Forecasting: Assessing Accuracy, Robustness,
	and Computational Efficiency},
	booktitle    = {{AIDS}},
	volume       = {3619},
	pages        = {12--22},
	year         = {2023}
}

@article{DBLP:journals/natmi/CuiA22,
	author       = {Peng Cui and Susan Athey},
	title        = {Stable learning establishes some common ground between causal inference
	and machine learning},
	journal      = {Nat. Mach. Intell.},
	volume       = {4},
	number       = {2},
	pages        = {110--115},
	year         = {2022}
}

@article{DBLP:journals/corr/abs-2505-17637,
	author       = {Yuting Huang and
	Ziquan Fang and
	Zhihao Zeng and
	Lu Chen and
	Yunjun Gao},
	title        = {Causal Spatio-Temporal Prediction: An Effective and Efficient Multi-Modal
	Approach},
	journal      = {CoRR},
	volume       = {abs/2505.17637},
	year         = {2025}
}

@inproceedings{DBLP:conf/icml/AhujaMWB23,
	author       = {Kartik Ahuja and
	Divyat Mahajan and
	Yixin Wang and
	Yoshua Bengio},
	title        = {Interventional Causal Representation Learning},
	booktitle    = {{ICML} },
	pages        = {372--407},
	year         = {2023}
}

@inproceedings{DBLP:conf/nips/AcarturkVST24,
	author       = {Emre Acart{\"{u}}rk and
	Burak Varici and
	Karthikeyan Shanmugam and
	Ali Tajer},
	title        = {Sample Complexity of Interventional Causal Representation Learning},
	booktitle    = {{NeurIPS }},
	year         = {2024}
}

@inproceedings{DBLP:conf/aaai/0016DLJW23,
	author       = {Yu Zhao and
	Pan Deng and
	Junting Liu and
	Xiaofeng Jia and
	Mulan Wang},
	title        = {Causal Conditional Hidden Markov Model for Multimodal Traffic Prediction},
	booktitle    = {{AAAI} },
	pages        = {4929--4936},
	year         = {2023}
}

@inproceedings{DBLP:conf/nips/XiaLWLWZZ23,
	author       = {Yutong Xia and
	Yuxuan Liang and
	Haomin Wen and
	Xu Liu and
	Kun Wang and
	Zhengyang Zhou and
	Roger Zimmermann},
	title        = {Deciphering Spatio-Temporal Graph Forecasting: {A} Causal Lens and
	Treatment},
	booktitle    = { {NeurIPS}},
	year         = {2023}
}

@inproceedings{DBLP:conf/iclr/WangWDZ0PZL024,
	author       = {Kun Wang and
	Hao Wu and
	Yifan Duan and
	Guibin Zhang and
	Kai Wang and
	Xiaojiang Peng and
	Yu Zheng and
	Yuxuan Liang and
	Yang Wang},
	title        = {NuwaDynamics: Discovering and Updating in Causal Spatio-Temporal Modeling},
	booktitle    = {{ICLR} },
	year         = {2024}
}

@inproceedings{DBLP:conf/nips/RajendranBASR24,
	author       = {Goutham Rajendran and
	Simon Buchholz and
	Bryon Aragam and
	Bernhard Sch{\"{o}}lkopf and
	Pradeep Ravikumar},
	title        = {From Causal to Concept-Based Representation Learning},
	booktitle    = {{NeurIPS}},
	year         = {2024}
}

@article{DBLP:journals/corr/abs-2602-06240,
	author       = {Yu Zhang and
	Sean Bin Yang and
	Arijit Khan and
	Cuneyt Gurcan Akcora},
	title        = {{ATEX-CF:} Attack-Informed Counterfactual Explanations for Graph Neural
	Networks},
	journal      = {CoRR},
	volume       = {abs/2602.06240},
	year         = {2026}
}

@inproceedings{DBLP:conf/cvpr/LvLLZLWL22,
	author       = {Fangrui Lv and
	Jian Liang and
	Shuang Li and
	Bin Zang and
	Chi Harold Liu and
	Ziteng Wang and
	Di Liu},
	title        = {Causality Inspired Representation Learning for Domain Generalization},
	booktitle    = {{CVPR} },
	pages        = {8036--8046},
	year         = {2022}
}

@article{DBLP:journals/csur/GongZYBLX25,
	author       = {Chang Gong and
	Chuzhe Zhang and
	Di Yao and
	Jingping Bi and
	Wenbin Li and
	Yongjun Xu},
	title        = {Causal Discovery from Temporal Data: An Overview and New Perspectives},
	journal      = {{ACM} Comput. Surv.},
	volume       = {57},
	number       = {4},
	pages        = {100:1--100:38},
	year         = {2025}
}

@inproceedings{DBLP:conf/icde/Li0GCJZZFB24,
	author       = {Wenbin Li and
	Di Yao and
	Chang Gong and
	Xiaokai Chu and
	Quanliang Jing and
	Xiaolei Zhou and
	Yuxuan Zhang and
	Yunxia Fan and
	Jingping Bi},
	title        = {CausalTAD: Causal Implicit Generative Model for Debiased Online Trajectory
	Anomaly Detection},
	booktitle    = { {ICDE} },
	pages        = {4477--4490},
	year         = {2024}
}

@inproceedings{DBLP:conf/icde/ChuLR023,
	author       = {Zhixuan Chu and
	Ruopeng Li and
	Stephen L. Rathbun and
	Sheng Li},
	title        = {Continual Causal Inference with Incremental Observational Data},
	booktitle    = { {ICDE} },
	pages        = {3430--3439},
	year         = {2023}
}

@inproceedings{DBLP:conf/icde/ZhouWWYDY25,
	author       = {Dehua Zhou and
	Bowei Wu and
	Ke Wang and
	Qifen Yang and
	Yuhui Deng and
	Siu{-}Ming Yiu},
	title        = {Intervention-Driven Correlation Reduction: {A} Data Generation Approach
	for Achieving Counterfactually Fair Predictors},
	booktitle    = { {ICDE} },
	pages        = {2066--2079},
	year         = {2025}
}

@article{DBLP:journals/pvldb/LiuXAW25,
	author       = {Jiaxiang Liu and
	Siyuan Xia and
	Daniel Alabi and
	Eugene Wu},
	title        = {Suna: Scalable Causal Confounder Discovery over Relational Data},
	journal      = {{PVLDB}},
	volume       = {18},
	number       = {11},
	pages        = {4158--4170},
	year         = {2025}
}

@inproceedings{DBLP:conf/ijcai/YangGHT021,
	author       = {Sean Bin Yang and Chenjuan Guo and Jilin Hu and Jian Tang and Bin Yang},
	title        = {Unsupervised Path Representation Learning with Curriculum Negative
	Sampling},
	booktitle    = { {IJCAI} },
	pages        = {3286--3292},
	year         = {2021}
}

@inproceedings{DBLP:conf/icde/YangGHYTJ22,
	author       = {Sean Bin Yang and
	Chenjuan Guo and
	Jilin Hu and
	Bin Yang and
	Jian Tang and
	Christian S. Jensen},
	title        = {Weakly-supervised Temporal Path Representation Learning with Contrastive
	Curriculum Learning},
	booktitle    = {{ICDE} },
	pages        = {2873--2885},
	year         = {2022}
}

@article{DBLP:journals/corr/abs-2601-08482,
	author       = {Chenxu Han and
	Sean Bin Yang and
	Jilin Hu},
	title        = {DiffMM: Efficient Method for Accurate Noisy and Sparse Trajectory
	Map Matching via One Step Diffusion},
	journal      = {CoRR},
	volume       = {abs/2601.08482},
	year         = {2026}
}

@inproceedings{DBLP:conf/kdd/YangHGYJ23,
	author       = {Sean Bin Yang and
	Jilin Hu and
	Chenjuan Guo and
	Bin Yang and
	Christian S. Jensen},
	title        = {LightPath: Lightweight and Scalable Path Representation Learning},
	booktitle    = {{KDD} },
	pages        = {2999--3010},
	year         = {2023}
}

@article{DBLP:conf/CIKM/TFM,
	title={Spatio-Temporal Trajectory Foundation Model-Recent Advances and Future Directions},
	author={Sean Bin Yang and Ying Sun and Yunyao Cheng and Yan Lin and Torp Kristian and Jilin Hu},
	journal={arXiv preprint arXiv:2511.20729},
	year={2025}
}

@article{DBLP:journals/tkde/YangGY22,
	author       = {Sean Bin Yang and
	Chenjuan Guo and
	Bin Yang},
	title        = {Context-Aware Path Ranking in Road Networks},
	journal      = {TKDE},
	volume       = {34},
	number       = {7},
	pages        = {3153--3168},
	year         = {2022}
}

@inproceedings{DBLP:conf/WWW/Path-LLM,
	author       = {Yongfu Wei and
	Yan Lin and
	Hongfan Gao and
	Ronghui Xu and
	Sean Bin Yang and
	Jilin Hu},
	title        = {Path-LLM: {A} Multi-Modal Path Representation Learning by Aligning
	and Fusing with Large Language Models},
	booktitle    = { {WWW} },
	pages        = {2289--2298},
	year         = {2025}
}

@inproceedings{DBLP:conf/KDD/MM-Path,
	author       = {Ronghui Xu and
	Hanyin Cheng and
	Chenjuan Guo and
	Hongfan Gao and
	Jilin Hu and
	Sean Bin Yang and
	Bin Yang},
	title        = {MM-Path: Multi-modal, Multi-granularity Path Representation Learning},
	booktitle    = {{KDD} },
	pages        = {1703--1714},
	year         = {2025}
}

@article{DBLP:journals/corr/abs-2203-16110,
	author       = {Sean Bin Yang and
	Chenjuan Guo and
	Jilin Hu and
	Bin Yang and
	Jian Tang and
	Christian S. Jensen},
	title        = {Weakly-supervised Temporal Path Representation Learning with Contrastive
	Curriculum Learning - Extended Version},
	journal      = {CoRR},
	volume       = {abs/2203.16110},
	year         = {2022},
}

@inproceedings{DBLP:conf/kdd/Yang26,
	author       = {Sean Bin Yang and Ying Sun and Jilin Hu and Zongyi Xu and Kristian Torp and Hua Lu and Bin Yang and Christian S. Jensen},
	title        = {REFINE: Trajectory Representation Learning via Closed-LoopTranscription},
	booktitle    = {{KDD}},
	year         = {2026}
}

@inproceedings{DBLP:conf/aaai/HanYH26,
	author       = {Chenxu Han and
	Sean Bin Yang and
	Jilin Hu},
	title        = {DiffMM: Efficient Method for Accurate Noisy and Sparse Trajectory
	Map Matching via One Step Diffusion},
	booktitle    = { {AAAI} },
	pages        = {14783--14791},
	year         = {2026}
}

@inproceedings{DBLP:conf/icde/Yang020,
	author       = {Sean Bin Yang and
	Bin Yang},
	title        = {Learning to Rank Paths in Spatial Networks},
	booktitle    = { {ICDE} },
	pages        = {2006--2009},
	year         = {2020}
}

@inproceedings{DBLP:conf/kdd/LiJZHC26,
	author       = {Zechen Li and
	Hongwei Jia and
	Kai Zhao and
	Weiming Huang and
	Meng Chen},
	title        = {Multi-View Urban Region Embedding via Commonality-Specificity Disentanglement},
	booktitle    = {{KDD} },
	pages        = {748--758},
	year         = {2026}
}

@inproceedings{DBLP:conf/mm/HazarikaZP20,
	author       = {Devamanyu Hazarika and
	Roger Zimmermann and
	Soujanya Poria},
	title        = {{MISA:} Modality-Invariant and -Specific Representations for Multimodal
	Sentiment Analysis},
	booktitle    = {{MM} },
	pages        = {1122--1131},
	year         = {2020}
}

@inproceedings{DBLP:conf/nips/BousmalisTSKE16,
	author       = {Konstantinos Bousmalis and
	George Trigeorgis and
	Nathan Silberman and
	Dilip Krishnan and
	Dumitru Erhan},
	title        = {Domain Separation Networks},
	booktitle    = {{NeurIPS}},
	pages        = {343--351},
	year         = {2016}
}

@inproceedings{DBLP:conf/ijcai/SunWHTH19,
	author       = {Yiwei Sun and
	Suhang Wang and
	Tsung{-}Yu Hsieh and
	Xianfeng Tang and
	Vasant G. Honavar},
	title        = {{MEGAN:} {A} Generative Adversarial Network for Multi-View Network
	Embedding},
	booktitle    = {{IJCAI} },
	pages        = {3527--3533},
	year         = {2019}
}

@article{DBLP:journals/pvldb/PanWZY0CGWTDZYZ23,
	author       = {Zhicheng Pan and
	Yihang Wang and
	Yingying Zhang and
	Sean Bin Yang and
	Yunyao Cheng and
	Peng Chen and
	Chenjuan Guo and
	Qingsong Wen and
	Xiduo Tian and
	Yunliang Dou and
	Zhiqiang Zhou and
	Chengcheng Yang and
	Aoying Zhou and
	Bin Yang},
	title        = {MagicScaler: Uncertainty-aware, Predictive Autoscaling},
	journal      = {Proc. {VLDB} Endow.},
	volume       = {16},
	number       = {12},
	pages        = {3808--3821},
	year         = {2023}
}

@article{DBLP:journals/corr/abs-1907-04028,
	author       = {Sean Bin Yang and
	Bin Yang},
	title        = {PathRank: {A} Multi-Task Learning Framework to Rank Paths in Spatial
	Networks},
	journal      = {CoRR},
	volume       = {abs/1907.04028},
	year         = {2019}
}

@article{DBLP:journals/corr/abs-REFINE,
	author       = {Sean Bin Yang and Ying Sun and Jilin Hu and Zongyi Xu and Kristian Torp and Hua Lu and Bin Yang and Christian S. Jensen},
	title        = {REFINE: Trajectory Representation Learning via Closed-Loop Transcription--Extended Version},
	journal      = {CoRR},
	volume       = {abs/2609.07206},
	year         = {2026}
}

@article{DBLP:journals/corr/abs-DGCPath,
	author       = {Sean Bin Yang and Hao Miao and Zongyi Xu and Jilin Hu and Xiangmeng Wang and Hua Lu and Bin Yang and Christian S. Jensen},
	title        = {DGCPath: Distribution-Aware Generative Contrastive Framework for Self-supervised Path Representation Learning -- Extended Version},
	journal      = {CoRR},
	volume       = {abs/2609.07316},
	year         = {2026}
}

\end{document}